%% file: iclr2027_conference.tex
\documentclass{article} % For LaTeX2e
\usepackage{iclr2027_conference,times}

\input{math_commands.tex}

\usepackage{hyperref}
\usepackage{url}

\usepackage{booktabs}
\usepackage[table]{xcolor}
\usepackage{subcaption}
\usepackage{xspace}
\usepackage{multirow}
\usepackage{graphicx}
\usepackage{wrapfig}

\usepackage{tikz}
\usetikzlibrary{tikzmark,decorations.pathreplacing}

\usepackage{algorithm}
\usepackage{algorithmicx}
\usepackage{algpseudocode}

\DeclareSymbolFont{boldoperators}{OT1}{cmr}{bx}{n}
\DeclareMathAccent{\bar}{\mathord}{boldoperators}{"16}

\title{Modeling Uncertainty Propagation in \\3D Morphable Models}

\iclrfinalcopy

\author{Megh Shukla \\
EPFL\\
\texttt{megh.shukla@epfl.ch} \\
\And
Edo Collins \\
Google \\
\texttt{edo.collins@google.com} \\
\And
Prashanth Chandran \\
Google \\
\texttt{prchandran@google.com}
\And
Karel Petránek \\
Google \\
\texttt{kpetranek@google.com}
\And
Sergio Orts-Escolano \\
Google \\
\texttt{sorts@google.com}
\And
Mathieu Salzmann \\
EPFL, Waabi\\
\texttt{mathieu.salzmann@epfl.ch}
\And
Alexandre Alahi \\
EPFL\\
\texttt{alexandre.alahi@epfl.ch} 
}

\newcommand{\methodnameshort}{CLUE\xspace}
\newcommand{\methodnamelong}{Closed-form Local Uncertainty Estimation\xspace}

\begin{document}

\maketitle

\input{abstract}
\input{introduction}
\input{related}
\input{CLUE}
\input{experiments}
\input{conclusion}
\input{statements}
\bibliography{iclr2027_conference}
\bibliographystyle{iclr2027_conference}

\newpage
\appendix
\include{appendix}

\end{document}

%% file: math_commands.tex
\usepackage{amsmath,amsfonts,bm}

\def\eqref#1{equation~\ref{#1}}
\def\1{\bm{1}}

\DeclareMathAlphabet{\mathsfit}{\encodingdefault}{\sfdefault}{m}{sl}
\SetMathAlphabet{\mathsfit}{bold}{\encodingdefault}{\sfdefault}{bx}{n}

%% file: abstract.tex
\begin{abstract}
We study the problem of mapping parameter-space variance of a 3D Morphable Model (3DMM) to per-vertex variance. Traditionally, per-vertex variance has been used to visually interpret the learnt parameter uncertainty. However, mapping parameter-space variance to per-vertex using sampling or error propagation is computationally expensive. We address this limitation with \textbf{C}losed-form \textbf{L}ocal \textbf{U}ncertainty \textbf{E}stimation (\methodnameshort), which transforms the parameter-space distribution of 3DMMs to per-vertex through a closed-form approximation. \methodnameshort emerges from applying the law of total variance to the 3DMM formulation and is modular, computationally inexpensive, and accurate. Subsequently, we study integrating \methodnameshort for in-the-wild human mesh recovery training and inference. We note that \methodnameshort uses the geometry of the mesh as an inductive bias when predicting the variance, leading to improved accuracy. Through extensive qualitative and quantitative experiments, we show that when used in training, \methodnameshort (1) outpeforms probabilistic and deterministic baselines on metrics evaluating joints, mesh and uncertainty, and (2) has the highest uncertainty-to-error spearman correlation, with applications including data mining and interpretable uncertainty. Finally, we share SAM 3D Body (CLUE) which augments the regressor with high resolution uncertainty maps. We make our code available at \url{https://github.com/vita-epfl/CLUE}
\end{abstract}

%% file: introduction.tex
\section{Introduction}

\begin{figure}[t]
    \centering
    \includegraphics[width=\linewidth]{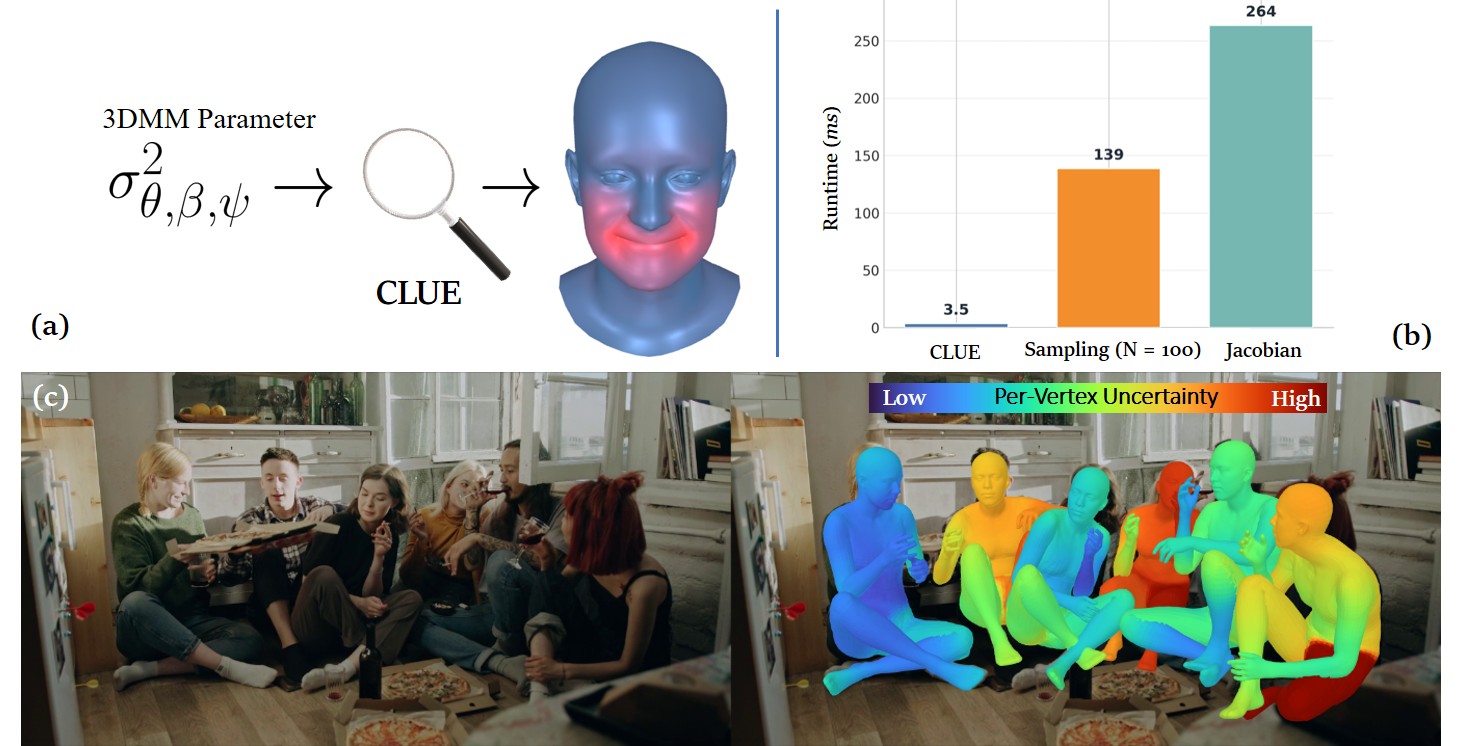}
    \caption{a) \textbf{\methodnameshort}: \textbf{C}losed-form \textbf{L}ocal \textbf{U}ncertainty \textbf{E}stimation is a novel transformation that maps parameter-space distribution to local, per-vertex distribution for 3D Morphable Models. b) \textbf{Accurate and Real-time}: \methodnameshort closely approximates sampling-based methods, while being \textbf{$\mathbf{\sim 40\times}$} faster. c) \textbf{In-the-wild}: We study in-the-wild training and inference, and share \textbf{\textit{SAM 3D Body (CLUE)}}, augmenting a frozen SAM 3D Body model with high-resolution uncertainty maps.}
    \label{fig:hook}
\end{figure}

3D Morphable Models (3DMMs) \cite{loper2023smpl, SMPL-X:2019, li2017learning} have been critical to the success of digital humans in real world applications. A popular application of 3DMMs is in-the-wild Human Mesh Recovery (HMR) \citep{kanazawa2018end, bogo2016keep, yang2026sam}, which reconstructs humans by estimating their 3DMM parameters from in-the-wild images. However, in-the-wild images often include factors such as occlusions, poor image quality and challenging poses. To address these uncertainties, methods have been developed to predict a distribution over the 3DMM parameters using the the Gaussian Negative Log-Likelihood (GNLL) and its variants \cite{sengupta2021probabilistic, sengupta2021hierarchical, dwivedi2024poco}. The parameter variance is then mapped to the mesh to obtain interpretable, spatially localized per-vertex uncertainty. However, mapping the parameter distribution onto the mesh is non-trivial.

Popular methods in human mesh regression \cite{sengupta2021probabilistic, sengupta2021hierarchical} compute per-vertex variances through repeated sampling. A large ($\approx100$) number of samples are drawn from the parameter distribution which are then used to obtain a set of meshes. The per-vertex variance is computed empirically from this set of meshes. This approach is, however, computationally intensive and is sensitive to the number of samples. In statistics, error propagation \cite{joint2008evaluation} is a powerful tool to map uncertainty across a smooth function. This typically involves computing the jacobian of the output \textit{w.r.t} the input and projecting the input variance through it. While error propagation provides a closed-form approximation of the per-vertex uncertainty, the technique is computationally expensive. Moreover, operators such as the Jacobian may lack compatibility with popular on-device machine learning frameworks such as XNNPACK \cite{XNNPACK}. 

To address this limitation, we present \methodnamelong (\methodnameshort), a novel formulation to map the parameter-space variance to per-vertex. \methodnameshort uses the law of total variance and error propagation to analyse the 3DMM formulation, disentangling variance from pose, shape and expression. As a result, \methodnameshort is interpretable not only by formulation but also visually. By making pragmatic assumptions, we drastically increase computational efficiency while retaining accuracy on par with linear error propagation. The reduced compute footprint coupled with accuracy of the transformation allow us to integrate \methodnameshort in training probabilistic in-the-wild human mesh regressors. Recent advances in optimizing under uncertainty \citep{pmlr-v235-shukla24a} highlight the use of inductive bias when modeling the predicted variance. In a similar vein, constrains the predicted parameter variance to explain the geometry of the 3DMM, resulting in improved accuracy of the HMR model. \methodnameshort is also agnostic to the architecture of the HMR model, allowing for a seamless integration in training and inference.

We study three popular 3DMMs: (1) FLAME \citep{li2017learning}, (2) SMPL-X \citep{SMPL-X:2019}, and (3) MHR \citep{ferguson2025mhr}, which are also representative of SMPL \cite{loper2023smpl} and STAR \cite{osman2020star}. Next, we perform two sets of experiments: (a) statistical experiments to evaluate \methodnameshort as a transformation from parameter to per-vertex distribution, and (b) real-world experiments to qualitatively and quantitatively evaluate the use of \methodnameshort in training HMR models. Our experiments confirm that \methodnameshort is accurate and computationally inexpensive, as well as outperforms architecture agnostic probabilistic baselines over the joints, mesh, negative log-likelihood and correlation metrics. Furthermore, we discuss applying CLUE for data mining and interpretable uncertainty for downstream decision making. Finally, we share SAM 3D Body \textit{(CLUE)}, which augments a frozen SAM 3D Body model with probabilistic vertices for uncertainty based reasoning. We share example visualizations in Figs.~\ref{fig:sam3dbody_clue_1}, ~\ref{fig:sam3dbody_clue_2}. We make our code publicly available at \url{https://github.com/vita-epfl/CLUE}

%% file: related.tex
\section{Related Work}

\textbf{3D Morphable Model}. 3DMMs are data-driven statistical models that represent different variations in human identity, expression and pose. They are typically constructed by applying Principal Component Analysis (PCA) to a registered collection of 3D meshes. These models represent geometry in rest-pose through a linear combination of shape, expression and pose-corrective bases. The rest-pose geometry can further be articulated through Linear Blend Skinning (LBS).

\begin{equation}
v_i' = \sum_{j \in \mathcal{J}} w_{i,j} \mathbf{G}_j(\theta, \mathbf{J}) \left( \bar{v}_i + B_{S,i}(\beta) + B_{E,i}(\psi) + \mathcal{K}_{P,i}(\theta) \right)
\label{eq:3DMM_simple}
\end{equation}

Following Eq. \ref{eq:3DMM_simple}, a 3DMM modifies a mean template $\bar{v}_i$ through a series of additive offsets. In this formulation, $B_S$, $B_E$, and $\mathcal{K}_P$ are the linear bases representing shape, expression, and pose correctives, respectively. To articulate the mesh, $\mathbf{G}_j$ represents the rigid transformation of each joint $J$, while $w_{i,j}$ denote the skinning weights that determine how these transformations influence each vertex. We kindly refer to FLAME \cite{li2017learning}, SMPL-X \cite{SMPL-X:2019}, Basel Face Model \citep{bfm09} for a more detailed description of how 3DMMs are built.

\textbf{Deterministic mesh regression.} There have been significant works which leverage 3DMMs for HMR such as \citet{kanazawa2018end}, SPIN \citep{kolotouros2019learning}, PARE \citep{kocabas2021pare}, Mesh Graphormer \citep{lin2021mesh}, CLIFF \citep{li2022cliff}, SAM 3D Body \citep{yang2026sam} and Fast SAM 3D Body \citep{yang2026fast}.  For faces, the classical 3DMM ~\cite{blanz2023morphable} underpins a large family of analysis-by-synthesis fitters, and deep regressors ~\cite{deng2019accurate} extend this to in-the-wild imagery. Yet, deterministic methods produce a point estimate and do not capture ambiguities in their prediction.

\textbf{Probabilistic approaches.}  Under strong occlusion, ProHMR \citep{kolotouros2021probabilistic} explicitly encodes the posterior as a multimodal distribution captured using normalizing flows. The use of normalizing flows is further refined in HuManiFlow \citep{sengupta2023humaniflow} and \citet{biggs20203d}. Diffusion and score-based generative models \citep{cho2023generative, stathopoulos2024score} explore its use in improved HMR. Similarly, for faces under occlusion, Diverse3DFace \citep{dey2022generating} and OFER \citep{selvaraju2025ofer} use generative modeling over the FLAME parameters. While multimodal approaches effectively capture diverse modes through sampling, they do not explicitly model the confidence of individual modes. Unimodal methods focus on modeling ambiguity due to occlusion, lighting, noise and partial visibility. One line of work models pose parameters which live on $SO(3)$, requiring that pose distributions must respect the rotation manifold \citep{sengupta2021hierarchical, dunkel2024normalizing, shetty2024propliks}. Methods such as \citet{egger2016occlusion,egger2018occlusion,li2023robust} focus on inference under occlusion and do not explicitly model the uncertainty. Along similar lines as this work, ~\citet{sengupta2021bmvc} uses error propagation, but the solution is applicable only in the rest pose with no rotations. When predicting Gaussians over both the shape \textit{and} pose parameters, \citet{sengupta2021probabilistic} defaults to using Monte Carlo (MC) sampling. To eliminate Monte Carlo sampling, POCO \citep{dwivedi2024poco} regresses uncertainty in a single forward pass. However, per-part uncertainty is derived heuristically and does not address the mapping of parameter distribution to the target distribution.

\textbf{Uncertainty Estimation.} Several approaches exist \citep{374138, gal2016dropout, lakshminarayanan2017simple, blundell2015weight} to quantify uncertainty in the prediction. Among them, \citet {NIPS2017_2650d608, 374138} jointly optimize the mean and variance of the prediction using GNLL. In our context, this translates to learning the 3DMM parameters and its variance. Transforming the parameter distribution to the mesh involves error propagation \citep{joint2008evaluation, schaefer2023glopro}, with the linear variant being  $\Sigma_y = J \Sigma_x J^T$, where $y = f(x)$. As we show later, computing the full Jacobian is computationally expensive. \citet{sengupta2021bmvc} uses error propagation but assumes that there is no rotation \textit{w.r.t} the canonical pose. TIC-TAC \citep{pmlr-v235-shukla24a} extends error propagation to general nonlinear networks, but remains computationally expensive and abstracts away the interpretability in the formulation. This leaves a gap in recent works on a closed-form transformation which is not only agnostic to any 3DMM formulation but is also accurate, fast and interpretable. With \methodnameshort, we propose an efficient solution for error propagation tailored for 3DMMs.

%% file: CLUE.tex
\section{\methodnamelong}

Let $X$ be the random variable denoting a vertex after applying additive offsets, and $Y$ be the random variable denoting the rigid transformation of the vertex. We derive \methodnameshort by asking what is $\text{Var}(Y)$? Our key insight is the law of total variance, which ties the uncertainty in $Y$ as a function of an explanatory variable $X$. We provide brief insights here and detail in the appendix (Sec.~\ref{app:derivation}).

\textbf{Assumptions.} We inherit our assumptions from the standard 3DMM formulation and related literature \citep{loper2023smpl, sengupta2021probabilistic, sengupta2021hierarchical, schaefer2023glopro, dwivedi2024poco}: the pose $\theta$, identity $\beta$, and expression $\psi$ parameters are independent. Next, we drop the pose correctives $\mathcal{K}_{P,i}(\theta)$ to allow for a simpler formulation, since the magnitude of change induced by the pose is much larger than those from pose correctives ($15 \text{ v.s. } 0.5$cm) \citep{osman2020star}. Dropping pose correctives allows for a simpler formulation; yet for completeness we discuss incorporating pose correctives in the appendix (Sec:~\ref{app:pose_correctives}).

\subsection{Derivation}

\textbf{Additive Offset Variance.} Let $X_i$ be the random variable denoting vertex $i$ after the additive offsets, $X_i = \bar{v}_i + \vec{\beta} B_{S,i} + \vec{\psi} B_{E,i}$, where the shape and expression parameters have variances $\sigma^2_\beta$ and $\sigma^2_\psi$. Because $X_i$ is linear in these parameters,  we have,
\begin{equation}
    \Sigma_{X_i} = \text{Var}[X_i] = B_{S,i}^T\, \mathrm{diag}(\sigma^2_\beta)\, B_{S,i} + B_{E,i}^T\, \mathrm{diag}(\sigma^2_\psi)\, B_{E,i} \;,
    \label{eq:var_blendshapes}
\end{equation}
We denote the entries of $\Sigma_{X_i}$ by $\Sigma^{(kl)}_{X_i}$, $k, l \in \{1, 2, 3\}$.

\textbf{Law of Total Variance.} The skinned vertex is denoted as $v'_i = \sum_{j \in \mathcal{J}} w_{i,j}\, \mathbf{G}_{j}(\theta, \mathbf{J})\, X_i$, where $\mathcal{J}$ is the set of joints and $w_{i,j}$ are the skinning weights. We have two sources of randomness, (1) the rest-pose vertex $X_i$, and (2) the pose $\theta$. The law of total variance splits $\text{Var}(v'_i)$ into the variance due to the pose, shape and expression:

\begin{equation}
    \text{Var}(v'_i) = \underbrace{\mathbb{E}_X\big[ \text{Var}_\theta(v'_i \mid X_i) \big]}_{\text{rotational variance}} + \underbrace{\text{Var}_X\big( \mathbb{E}_\theta[v'_i \mid X_i] \big)}_{\text{shape and expression variance}} \,.
    \label{eq:ltv}
\end{equation}

Intuitively, the rotational variance quantifies how much a vertex moves when the pose varies for an expected body shape. The shape and expression variance quantifies the movement due to the shape and expression given the expected pose.
 
\textbf{Rotational Variance.} Let us consider the rest-pose vertices deterministic and let the pose vary. Because $\mathbf{G}_j$ is composed along the kinematic chain, a small rotation $\delta\omega_a$ of joint $a$ swings all descendent joints $\mathcal{D}(a)$ rigidly about the mean posed location $\bar{\mathbf{t}}_a$ of that joint. A point at $x$ thus moves by $\delta\omega_a \times (x - \bar{\mathbf{t}}_a)$, a small angle times a lever arm. For the skinned vertex, the lever arm is averaged over the joints that carry it and lie below $a$, and the displacements of all joints add:
\[
    v'_i \;\approx\; \sum_{j \in \mathcal{J}} w_{i,j}\, \bar{\mathbf{G}}_j X_i \;+\; \sum_{a \in \mathcal{J}} \delta\omega_a \times \boldsymbol{\ell}_{i,a} \,,
    \qquad
    \boldsymbol{\ell}_{i,a} = \sum_{j \in \mathcal{D}(a)} w_{i,j}\, \big( \bar{\mathbf{G}}_j X_i - \bar{\mathbf{t}}_a \big) \,.
\]
A joint contributes only if a descendent of the joint skins vertex $i$. For instance uncertainty travels down the chain, from the shoulder to the hand, but not across it. The lever arm grows with the distance from the joint, so the same angular uncertainty at the shoulder produces a larger positional uncertainty at the fingertip than at the elbow. For each joint, we have a Gaussian on its axis-angle coordinates $\omega_a$ in the axes of the parent joint, with independent per-axis variances $\sigma^2_{\theta_a, x}, \sigma^2_{\theta_a, y}, \sigma^2_{\theta_a, z}$.An increment $\varepsilon_a$ of these coordinates is the small rotation $\delta_a = J_l(\bar{\omega}_a)\, \varepsilon_a$ composed with the mean rotation, where $J_l$ is the left Jacobian of $SO(3)$. In world axes, the covariance of the perturbation is therefore $\Sigma_{\theta_a} = \bar{\mathbf{R}}_{p(a)}\, J_l(\bar{\omega}_a)\, \text{diag}(\sigma^2_{\theta_a, x}, \sigma^2_{\theta_a, y}, \sigma^2_{\theta_a, z})\, J_l(\bar{\omega}_a)^T\, \bar{\mathbf{R}}_{p(a)}^T$, where $\bar{\mathbf{R}}_{p(a)}$ is the mean global rotation of the parent. The variance of the displacement comes from the cross product. Intuitively, a rotation about $y$ moves the point in $x$ by its $z$ lever arm and in $z$ by its $x$ lever arm, and likewise for the other axes. Evaluating the lever arm at the mean vertex, with components $(\ell^{(x)}_{i,a}, \ell^{(y)}_{i,a}, \ell^{(z)}_{i,a})$, and writing $\Sigma^{(pq)}_{\theta_a}$ for the entries of $\Sigma_{\theta_a}$, the rotational variance is
\begin{equation}
    \mathbb{E}_X \big[\text{Var}_{\theta}(v'_i \, | \, X_i)\big] \approx \sum_{a \in \mathcal{J}} \begin{bmatrix}
\Sigma^{(yy)}_{\theta_a} \big(\ell^{(z)}_{i,a}\big)^2 + \Sigma^{(zz)}_{\theta_a} \big(\ell^{(y)}_{i,a}\big)^2 - 2\, \Sigma^{(yz)}_{\theta_a}\, \ell^{(y)}_{i,a}\, \ell^{(z)}_{i,a} \\
\Sigma^{(xx)}_{\theta_a} \big(\ell^{(z)}_{i,a}\big)^2 + \Sigma^{(zz)}_{\theta_a} \big(\ell^{(x)}_{i,a}\big)^2 - 2\, \Sigma^{(xz)}_{\theta_a}\, \ell^{(x)}_{i,a}\, \ell^{(z)}_{i,a} \\
\Sigma^{(xx)}_{\theta_a} \big(\ell^{(y)}_{i,a}\big)^2 + \Sigma^{(yy)}_{\theta_a} \big(\ell^{(x)}_{i,a}\big)^2 - 2\, \Sigma^{(xy)}_{\theta_a}\, \ell^{(x)}_{i,a}\, \ell^{(y)}_{i,a}
\end{bmatrix}.
\label{eq:final_rotation}
\end{equation}

 \begin{figure}[t]
    \centering
    \includegraphics[width=\linewidth]{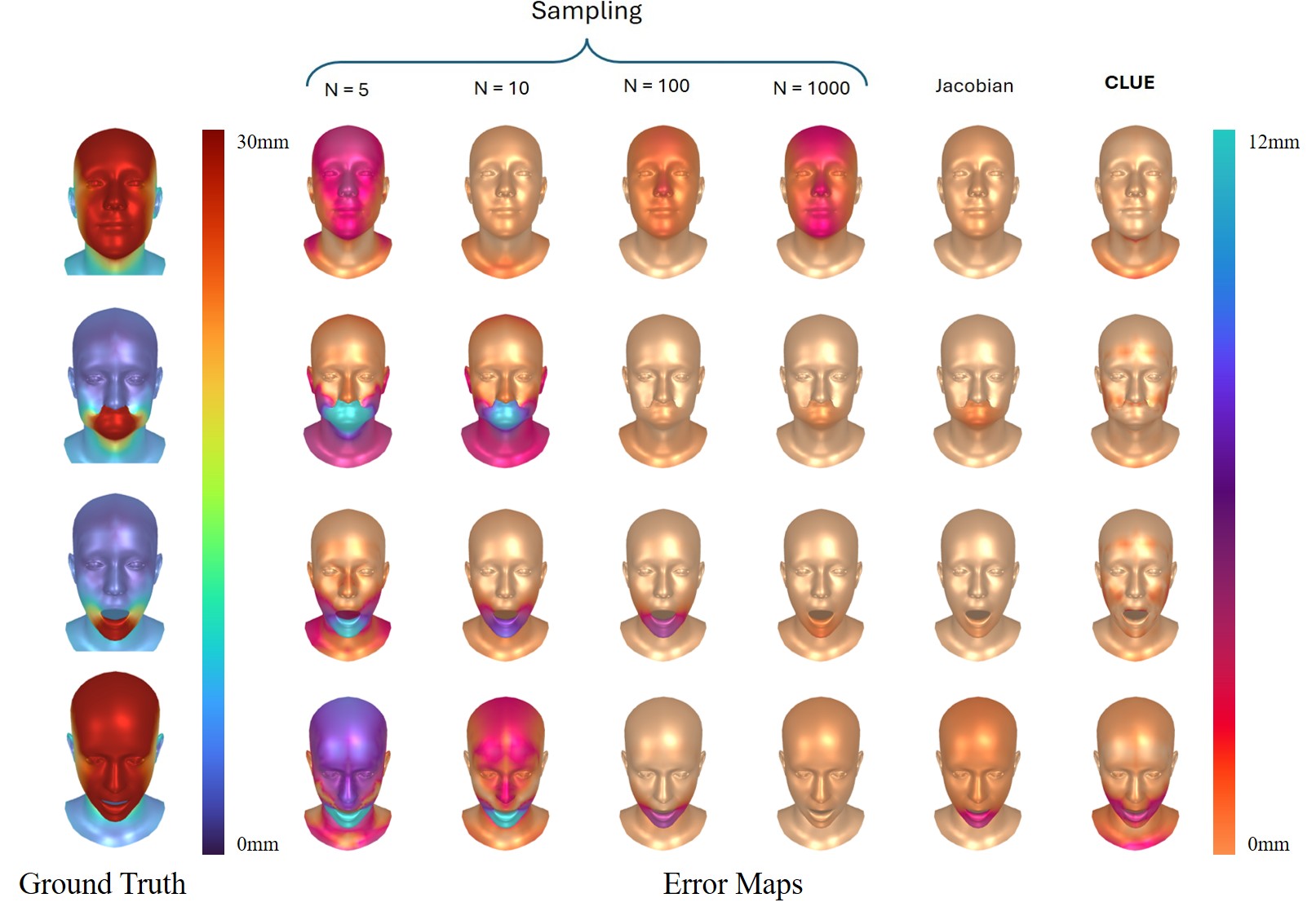}
    \caption{\textbf{FLAME Error Map}: We visualize the per-vertex error between the ground truth variance (computed with $N=10,000$ samples) and the estimated variance from different baselines. To do so, we synthesize different faces by setting randomly chosen parameters and their associated variance to non-zero. We observe that sampling with N=1000, Jacobian based error propagation and \methodnameshort have low errors. Unlike sampling, Jacobian and \methodnameshort may \textit{overestimate} the variance due to first order assumptions. Additional visualizations in Fig.~\ref{fig:error_maps_app}.}
    \label{fig:flame_error}
\end{figure}

\textbf{Shape and Expression Variance.} Let us now fix the pose at its mean and let the rest-pose offsets vary. The expected posed vertex is $\mathbb{E}_{\theta}[v'_i \, | \, X_i] = \big(\sum_{j \in \mathcal{J}} w_{i,j}\, \bar{\mathbf{G}}_j\big) X_i$. The offset covariance of Eq.~\ref{eq:var_blendshapes} is simply rotated into the posed body by the vertex's blended rotation $\mathbf{R}^{\mathrm{eff}}_i = \sum_{j \in \mathcal{J}} w_{i,j}\, \bar{\mathbf{R}}_j$. Using our identity $\text{Var}\big[ AZ \big] = A \Sigma_Z A^T$ and keeping the diagonal of $\mathbf{R}^{\mathrm{eff}}_i\, \Sigma_{X_i}\, (\mathbf{R}^{\mathrm{eff}}_i)^T$, we have
\begin{equation}
    \text{Var}_{X}\big(\mathbb{E}_{\theta}[v'_i \, | \, X_i]\big) = \begin{bmatrix}
\sum_{k,l=1}^{3} \mathbf{R}^{\mathrm{eff}}_{i, 1, k}\, \mathbf{R}^{\mathrm{eff}}_{i, 1, l}\, \Sigma^{(kl)}_{X_i} \\
\sum_{k,l=1}^{3} \mathbf{R}^{\mathrm{eff}}_{i, 2, k}\, \mathbf{R}^{\mathrm{eff}}_{i, 2, l}\, \Sigma^{(kl)}_{X_i} \\
\sum_{k,l=1}^{3} \mathbf{R}^{\mathrm{eff}}_{i, 3, k}\, \mathbf{R}^{\mathrm{eff}}_{i, 3, l}\, \Sigma^{(kl)}_{X_i}
\end{bmatrix}.
\label{eq:final_blendshape}
\end{equation}

Substituting Eq.~\ref{eq:final_rotation} and Eq.~\ref{eq:final_blendshape} into Eq.~\ref{eq:ltv} gives the per-vertex variance of the 3DMM formulation.
 
In contrast to first order error propagation, \methodnameshort leverages the 3DMM formulation without needing to compute the full Jacobian. The per-vertex variance cleanly disentangles the different sources of uncertainty and combines them through a simple summation. As a result, estimating this quantity is inexpensive in terms of compute time and memory while remaining accurate. We provide pseudo-code in the appendix (Sec.~\ref{app:pseudocode}) as well as the code implementation for FLAME, SMPL-X and MHR in the repository.

\begin{figure}[t]
    \centering
    \includegraphics[width=\linewidth]{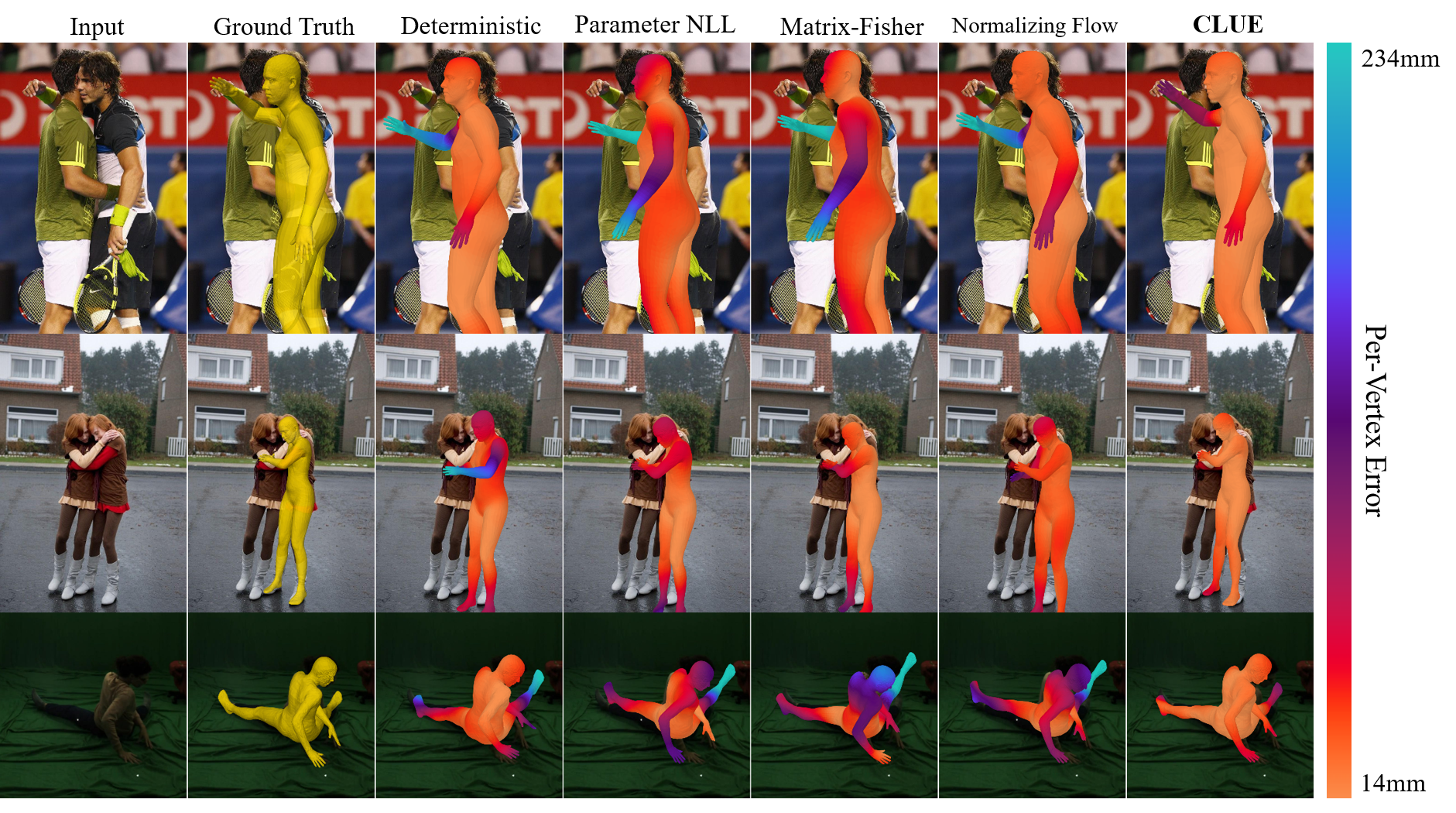}
    \caption{\textbf{Per-vertex error maps.} Each row shows the input, the
reference mesh (yellow), and the predictions of different baselines, coloured by the per-vertex error. The largest baseline errors sit on limbs that are occluded or having complex poses. \methodnameshort uses the mesh as an inductive bias when predicting the variance, allowing it to have lower errors when dealing with ambiguity.  Additional visualizations are shown in Fig.~\ref{fig:itw_errormaps_appendix}.}
    \label{fig:itw_error}
\end{figure}

\subsection{Training: CLUE with SAM 3D Body} 

Methods such as \citet{sengupta2021probabilistic, dwivedi2024poco} which rely on the negative log-likelihood to train the variance may suffer from suboptimal convergence. This is because the variance head lacks direct supervision and may take on arbitrary values that minimize the objective sub-optimally \citep{seitzerpitfalls, stirn2023faithful, immer2023effective, pmlr-v235-shukla24a, shukla2025towards}. In \citet{pmlr-v235-shukla24a}, a solution was proposed which used inductive bias when modeling the predicted variance. \methodnameshort presents one such inductive bias by incorporating the geometry of the 3DMM when learning the parameter variance. 

\textbf{Variance Head.} \methodnameshort is architecture agnostic, hence we make no changes to the architecture of SAM 3D Body. We use minimal additional parameters to create a variance head. The variance head takes as input the pose token from SAM 3D Body which is also consumed by the MHR head. In addition, the variance head also takes in the mean parameters output by the MHR head. Both the pose token and the mean parameters are concatenated before being fed to a two layer feedforward network with one hidden layer of width 1024 and ReLU activation. The output of this network is the variance for the shape and pose parameters, constrained by sigmoid activation but scaled by learnable scalars. These predicted variances are consumed by \methodnameshort. Note that all our probabilistic baselines share the same variance head architecture.

\textbf{Losses.} By defining a transformation between the parameter, vertex and joints space, \methodnameshort defines an equivalency between them. This implies that the parameter variance can be mapped to vertex and joint space without any additional learnable parameters. This is particularly useful because while the joints are traditionally inferred from the vertices, \methodnameshort allows inferring the joints as well as their variances from the vertices trivially. Thus, this enables \methodnameshort to use the joint-space negative log-likelihood (joints-NLL), replacing the traditional deterministic joint-space loss. Therefore, the geometry of the 3DMM guides the optimization of the parameter variance, which has to explain both: the parameter-NLL as well as joints-NLL. To compute joints-NLL, we assume that neighboring vertices are correlated. Hence the joint standard deviation is the weighted average of the vertex standard devation, $\sigma_{\text{joint}} = \sum_v w_v\,\sigma_{\mathbf{v}}$. In contrast, assuming independence meant that the joint variance  was underestimated by $5$ to $13\times$.

\subsection{Limitations}

\methodnameshort like jacobian based error propagation is a linear approximation, and overestimates the per-vertex variance when used as a statistical transformation. This is particularly visible for extreme body poses as shown in Fig.~\ref{fig:flame_error}. Furthermore, \methodnameshort leverages the structure of the 3DMM and does not extend to neural shape representation approaches. Finally, \methodnameshort makes several approximations to drastically reduce the number of computations needed without sacrificing accuracy, such as first-order analysis and dropping covariances between the vertices.

%% file: experiments.tex
\section{Experiments}

\begin{figure}[t]
    \centering
    \includegraphics[width=\linewidth]{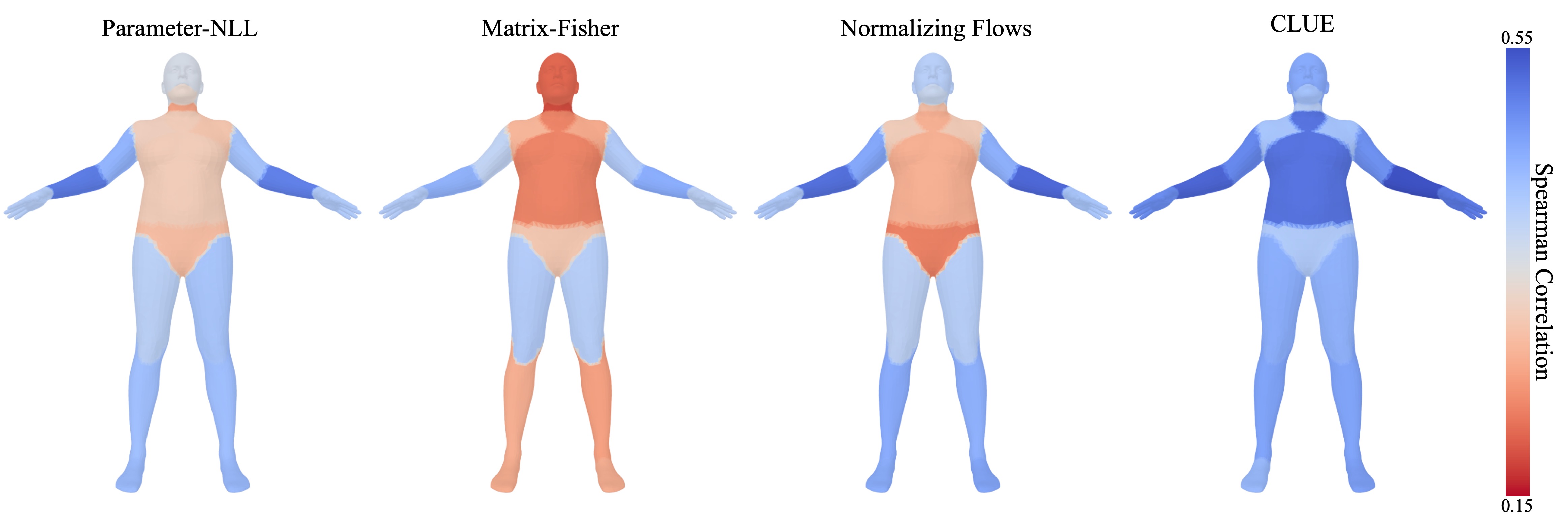}
    \caption{\textbf{Correlations across vertex groups.} We compute the Spearman correlation between the predicted per-vertex variance and the rigidly aligned vertex error for each of the 19 MHR vertex groups. We observe that \methodnameshort has a higher spearman correlation for each of the vertex groups, enabling applications in data mining and uncertainty based decision making.}
    \label{fig:group_corr}
\end{figure}

We perform two sets of experiments, designed to evaluate the accuracy of CLUE as a transformation from parameter-space to vertex-space, as well as the role of CLUE to improve optimization when training in-the-wild human mesh recovery models.

\subsection{Statistical Evaluation}

\setlength{\columnsep}{14pt}
\begin{wraptable}{r}{0.55\textwidth}
\vspace{-12pt}
\centering
\footnotesize
\setlength{\tabcolsep}{4pt}
\renewcommand{\arraystretch}{1.05}
\caption{\textbf{FLAME compute.} Median of 10 trials after 3 warm-ups. Additional results in Tab.~\ref{tab:compute-cost}.}
\label{tab:flame-cost}
\begin{tabular}{l rrr}
\toprule
\textbf{Method} & Time (ms) & Mem.\ (MB) & GFLOPs \\
\midrule
Sampling $N{=}100$  & 138.7  & 14.2  & 1.43 \\
Sampling $N{=}1000$ & 1379.4 & 140.9 & 14.26 \\
Jacobian            & 226.4  & 410.4 & 6.01 \\
\midrule
\rowcolor{gray!12}
\textbf{CLUE (ours)} & \textbf{3.4} & \textbf{4.3} & \textbf{0.039} \\
\bottomrule
\end{tabular}
\vspace{-2pt}
\end{wraptable}
The first set of experiments evaluates CLUE as a transformation from parameter-space to the space of meshes, studying accuracy and compute cost. These experiments rely on synthesizing parameter variance, and comparing how \methodnameshort propagates these variances in comparison to sampling with $N =\{5, 10, 100, 1000\}$ and error propagation. We perform these studies across FLAME, SMPL-X and Momentum Human Rig (MHR).

\textbf{Accuracy} (Fig. \ref{fig:flame_error}). We compute the \textit{ground truth} variance by simulating parameter variance, drawing $N = 10,000$ samples, and computing per-vertex variance empirically from the resulting meshes. For each baseline, we compute the absolute difference between its variance and that of the ground truth. We visualize these as error maps by randomly setting different parameter groups and their associated variance. For sampling based methods, as the value of $N$ increases, there is a drop in the error across all parameter groups. We also observe that $N = 1000$ and jacobian error propagation closely approximates the ground truth. While \methodnameshort similarly has low errors across different parameter groups, we highlight the intricate pattern on the forehead, which also comes up in the error map for \methodnameshort. We hypothesize that since \methodnameshort does not model pose correctives, it does not capture intricate changes in the mesh as the jaw joint moves.

\textbf{Compute}. We evaluate on runtime, peak incremental memory, and GFLOPs. We implement the Jacobian in forward mode in PyTorch \citep{paszke2019pytorch}. We use the CPU for computing the transformations to reflect on-device computing constraints, where the GPU would be prioritized for batched inferring from the incoming data stream. \methodnameshort has the fastest runtime, with error propagation via the Jacobian consuming the largest memory. \methodnameshort requires $\approx$ 4MB of additional memory because the tensors needed to compute the variance are already on-device. For the larger SMPL-X and MHR rigs (Tab.~\ref{tab:compute-cost}), our implementation exploits the sparsity of the kinematic tree, computing lever arms only for the joints that skin a vertex and their ancestors. The low computational costs as well as accuracy of the transformation makes \methodnameshort suitable for on-device uncertainty estimation.

\subsection{Probabilistic In-the-wild Human Mesh Recovery}

\begin{figure}[t]
    \centering
    \includegraphics[width=\linewidth]{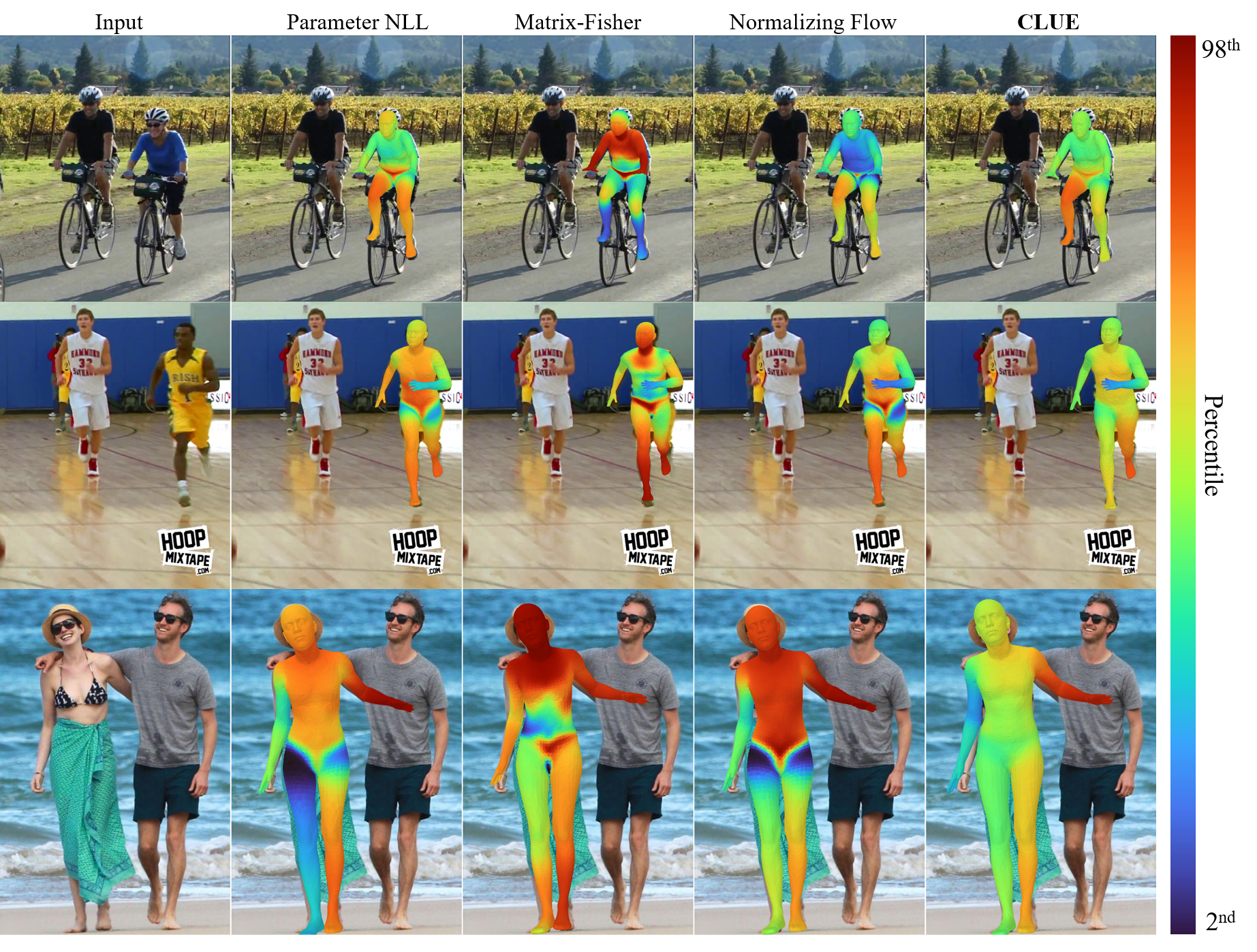}
    \caption{\textbf{Uncertainty based data-mining.} Per-vertex variance is computed empirically by sampling N=100 for all other baselines, and using \methodnameshort for itself. Similar to spearman correlation, for each method we normalize a vertex's uncertainty based on its per-vertex percentile for that method. These images were retrieved algorithmically by querying for images with the highest uncertainty discrepancy between a limb and the body. In contrast to CLUE, the baselines' uncertainty on the same images is body-wide or diffused and lacks fine-grained resolution. Additional visualizations are shown in Fig.~\ref{fig:datamining_appendix}.}
    \label{fig:datamining}
\end{figure}

The second set of experiments studies CLUE when used to supervise the training of unimodal in-the-wild human mesh recovery models such as SAM 3D Body. We compare our method with deterministic training, and architecture-agnostic probabilistic losses for uncertainty such as 3DMM Parameter-NLL, Matrix Fisher, and Normalizing Flows \citep{sengupta2021probabilistic, sengupta2021hierarchical, dwivedi2024poco, kolotouros2021probabilistic}. We randomly initialize the SAM 3D Body \cite{yang2026sam} architecture with frozen DINOv3 \cite{dinov3} backbone as our HMR model. We describe the losses, datasets and training process in the appendix (Sec:~\ref{app:training}).

\subsubsection{Quantitative Study}

\input{tables/3dpw_main}

\textbf{3DPW and EMDB.} (Table~\ref{tab:3dpw_emdb}) We observe that all probabilistic losses outperform the deterministic loss. By introducing inductive bias when modeling the mesh, \methodnameshort results in improved optimization and is the strongest baseline over all metrics. We also report the evaluation of SAM 3D Body on our pipeline which we recall is trained on a significantly larger dataset. Our numbers differ from those published due to possible differences with the SAM 3D Body evaluation pipeline which is \href{https://github.com/facebookresearch/sam-3d-body/issues/111}{not public}. While \methodnameshort is competitive with SAM 3D Body on 3DPW, it falls short on EMDB. We hypothesize that probabilistic training can reason better when dealing with ambiguities, but cannot model poses that are out-of-distribution accurately. This suggests a strategy of probabilistic modeling coupled with a diverse dataset.

\input{tables/pGT_compact}

\textbf{Pseudo Ground Truth datasets.} Table~\ref{tab:pgt-compact} scores each baseline against the released SAM 3D Body as ground truth on five image datasets without motion capture, in MHR space. By having the prediction and ground truth in MHR space, we can compute uncertainty related metrics. \methodnameshort is the only probabilistic objective that improves every group of metrics. Figure~\ref{fig:group_corr} visualizes the spearman correlation between the vertex error after rigid alignment of the ground truth and predicted mesh. We observe that \methodnameshort has the highest correlation across all groups.

\subsubsection{Qualitative Study}

\textbf{Error Maps.} Figure~\ref{fig:itw_error} colours each prediction by its per-vertex error against the reference mesh, on one shared scale. We observe two patterns. First, the baselines struggle with limbs that are occluded or have complex poses. \methodnameshort is resistant to occlusion due to the inductive bias of the mesh in modeling the variance and allows for probabilistic reasoning under occlusion. Second, in complex body poses such as floor exercises and splits, the parameter-space baselines spread error over the torso and head as well, whereas \methodnameshort confines the error to the extremities.

\textbf{Uncertainty based data mining.}  Since \methodnameshort predicts fine-grained uncertainty maps (evidenced by high spearman correlations), we can answer questions that parameter-space uncertainty cannot: \emph{which} vertex groups of the body is the model unsure about? For this study, we look for a limb whose uncertainty stands out from the rest of the body. However, raw uncertainty values are not comparable as they differ in scale between methods and vertices. We therefore convert each vertex's uncertainty into a percentile, i.e. how unusual that value is for that vertex and that method, measured against the SA-1B training set. We also recall that per-vertex variances are computed empirically for all baselines except \methodnameshort. In Fig.~\ref{fig:datamining} we observe that CLUE correctly distinguishes vertex groups which are occluded or undergo motion blur, while the rest of the body remains confident. The baselines saturate uncertainty across the whole body. While they can say that an image is hard, they cannot localize where.

\begin{figure}[t]
    \centering
    \includegraphics[width=\linewidth]{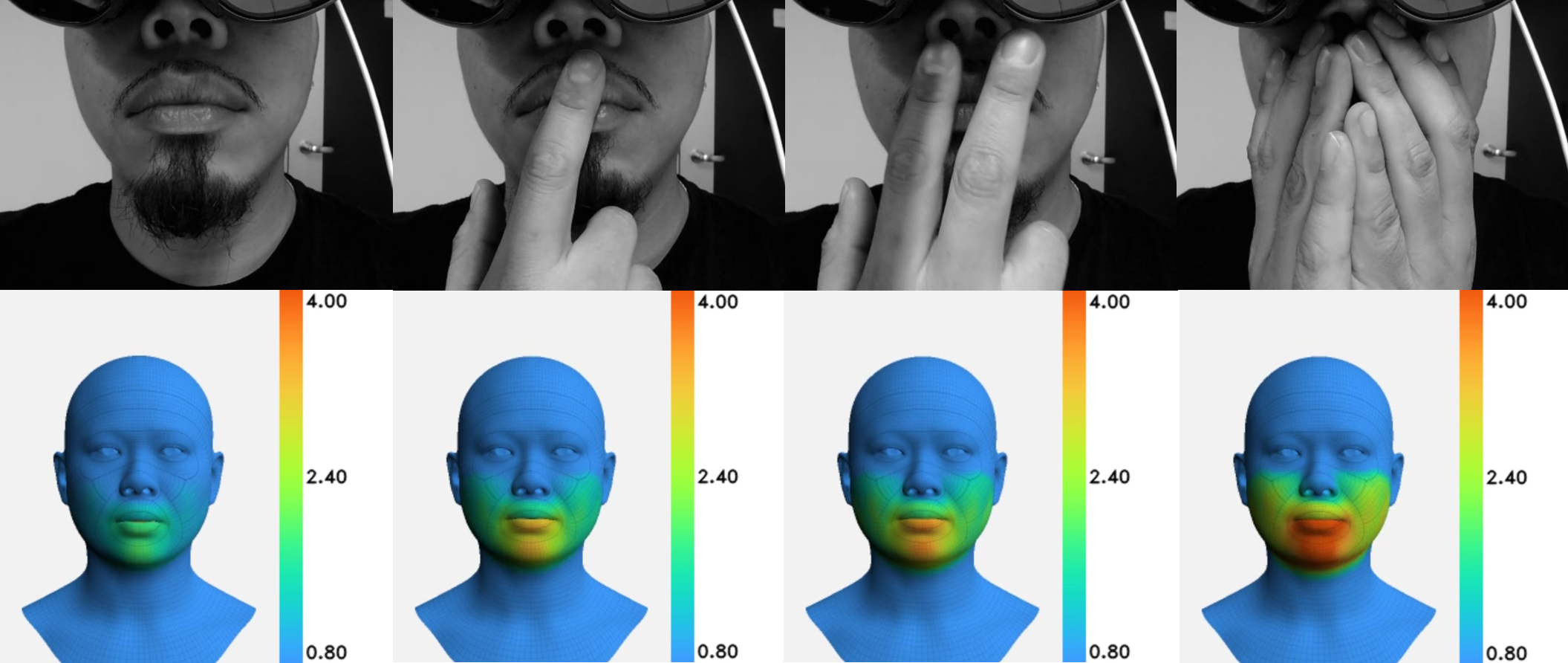}
    \caption{\textbf{Uncertainty for downstream decision making.} Top: frames from a headset's egocentric camera with the
    mouth progressively occluded. Bottom: the tracked 3DMM per-vertex standard deviation (mm). The uncertainty concentrates on the occluded region, rising monotonically with the degree of occlusion. With this signal, a system can suspend or down-weight vision-based face tracking during occlusion, or defer to another modality, instead of tracking blindly.}
    \label{fig:ft}
\end{figure}

\textbf{Uncertainty based decision making.} We present a use-case in egocentric face tracking. This study uses an internal dataset and GNM \cite{ploumpis2026gnmhead} as the 3DMM. The goal is to regress a 3D mesh that reproduces the lower-face expression of a person from egocentric viewpoints. This setup is typical of VR-headsets such as Apple Vision Pro, Meta Quest Pro, and Galaxy XR, which further impose strict runtime and memory restrictions to enable realtime on-device operation. We employ a similar training protocol as described earlier to train the probabilistic regressor. We observe in Fig.~\ref{fig:ft} that as the degree of occlusion increases, the uncertainty is localized and scaled correctly. Given this uncertainty signal, a system no longer has to track blindly through an occlusion: it can pause face tracking, lower its weight in the fusion, or hand over to another sensor until the face is visible again.

%% file: tables/3dpw_main.tex
\begin{table}[t]
\caption{\textbf{3DPW and EMDB (MoCap SMPL).} All metrics in
mm ($\downarrow$). The first row measures SAM 3D Body on our evaluation pipeline. The other rows compare deterministic and probabilistic losses under identical training and evaluation protocol.}
\label{tab:3dpw_emdb}
\begin{center}
\small
\setlength{\tabcolsep}{5pt}
\renewcommand{\arraystretch}{1.15}
\resizebox{\linewidth}{!}{%
\begin{tabular}{l c cccc cccc}
\toprule
& & \multicolumn{4}{c}{\textsc{3DPW}} & \multicolumn{4}{c}{\textsc{EMDB}} \\
\cmidrule(lr){3-6} \cmidrule(l){7-10}
\textbf{Method} & \#Training Images & MPJPE & PA-MPJPE & PVE & PA-PVE & MPJPE & PA-MPJPE & PVE & PA-PVE \\
\midrule
\textcolor{gray}{SAM 3D Body (released)} & \textcolor{gray}{3.8M (w/ multi-view)} & \textcolor{gray}{56.0} & \textcolor{gray}{35.1} & \textcolor{gray}{67.1} & \textcolor{gray}{47.2} & \textcolor{gray}{63.2} & \textcolor{gray}{38.2} & \textcolor{gray}{74.7} & \textcolor{gray}{47.1} \\
%\textcolor{gray}{CLUE (ours), @100k steps} & \textcolor{gray}{1.2M (single-view)} & \textcolor{gray}{55.8} & \textcolor{gray}{34.5} & \textcolor{gray}{66.4} & \textcolor{gray}{46.7} & \textcolor{gray}{64.9} & \textcolor{gray}{39.7} & \textcolor{gray}{77.0} & \textcolor{gray}{49.0} \\
\arrayrulecolor{black!15}
\midrule
\tikzmark{top}Deterministic Training & 1.2M (single-view) & 62.7 & 37.2 & 72.9 & 49.7 & 71.9 & 43.4 & 85.5 & 52.6 \\
Parameter NLL & '' & 57.9 & 35.5 & 68.9 & 48.3 & 66.4 & 41.2 & 79.5 & \textbf{50.9} \\
Matrix-Fisher  & '' & 58.0 & 35.7 & 68.8 & 48.1 & 68.8 & 42.5 & 81.7 & 51.8 \\
Normalizing Flow      & '' & 58.3 & 35.6 & 68.9 & 48.0 & 68.0 & 41.4 & 80.7 & \textbf{50.8} \\
\arrayrulecolor{black}
\midrule
\rowcolor{gray!12}
\tikzmark{bot}\textbf{CLUE} & '' & \textbf{56.1} & \textbf{34.9} & \textbf{67.0} & \textbf{47.2} & \textbf{65.4} & \textbf{40.8} & \textbf{77.9} & \textbf{50.7} \\
\bottomrule
\end{tabular}%
\begin{tikzpicture}[overlay, remember picture]
  \draw[decorate, decoration={brace, amplitude=5pt}, thick]
    ([shift={(-\tabcolsep-4pt,-3.8pt)}]pic cs:bot) -- ([shift={(-\tabcolsep-4pt,8.9pt)}]pic cs:top)
    node[midway, shift={(-8pt,0)}, rotate=90, anchor=south] {\footnotesize Same training recipe};
\end{tikzpicture}%
}
\end{center}
\end{table}

%% file: tables/pGT_compact.tex
\begin{table}[t]
\caption{\textbf{Pseudo-ground-truth evaluation}, \textit{averaged} over five
in-the-wild datasets (H36M, MPI-INF-3DHP, PoseTrack, CrowdPose,
OCHuman): joint and surface error (mm, $\downarrow$), vertex NLL
($\downarrow$), Pearson correlation between predicted uncertainty and error ($\uparrow$), and the area under the sparsification error curve (AUSE, $\downarrow$; per person, ranking by the mean predicted vertex sigma). Per-dataset results are in the appendix (Tab.~\ref{tab:pgt-main}).}
\label{tab:pgt-compact}
\centering
\small
\setlength{\tabcolsep}{6pt}
\renewcommand{\arraystretch}{1.15}
\resizebox{0.8\linewidth}{!}{%
\begin{tabular}{l cc cc ccc}
\toprule
& \multicolumn{2}{c}{\textsc{Joints}} & \multicolumn{2}{c}{\textsc{Surface}} & \multicolumn{3}{c}{\textsc{Uncertainty}} \\
\cmidrule(lr){2-3} \cmidrule(lr){4-5} \cmidrule(l){6-8}
\textbf{Method} & MPJPE & PA-MPJPE & PVE & PA-PVE & vNLL & $\rho$ & AUSE \\
\midrule
Deterministic    & 31.6 & 19.7 & 35.4 & 24.2 & -- & -- & -- \\
Param-NLL        & 32.0 & 18.5 & 35.0 & 22.3 & 2.35 & 0.68 & 0.078 \\
Matrix-Fisher    & 35.9 & 20.5 & 38.9 & 24.4 & 2.74 & 0.56 & 0.129 \\
Normalizing Flow & 32.0 & 18.1 & 35.3 & 22.2 & 3.08 & 0.69 & 0.077 \\
\arrayrulecolor{black}
\midrule
\rowcolor{gray!12}
\textbf{CLUE} & \textbf{27.5} & \textbf{17.7} & \textbf{30.4} & \textbf{21.7} & \textbf{2.13} & \textbf{0.75} & \textbf{0.067} \\
\bottomrule
\end{tabular}%
}
\end{table}

%% file: conclusion.tex
\section{Conclusion}
We proposed \methodnameshort, a novel formulation which transforms the parameter-space distribution of a 3DMM onto the mesh in closed-form. Through statistical experiments across multiple 3DMMs, we showed that \methodnameshort is an accurate transformation while being computationally efficient. We also noted that in comparison to a direct application of linear error propagation, \methodnameshort is interpretable in terms of formulation and visually. We studied integrating \methodnameshort for in-the-wild human mesh recovery training and inference. Because \methodnameshort carries the predicted parameter variance through the geometry of the 3DMM, it serves as an inductive bias to train the variance head. Subsequently, we show that \methodnameshort outperforms deterministic training as well as architecture agnostic unimodal uncertainty losses. Next, we analyze the variance learnt by \methodnameshort, showing that it has the highest spearman correlation for all vertex groups, and discuss applications of the learnt uncertainty in data mining and downstream decision making. Finally, we share SAM 3D Body (\methodnameshort) to make per-vertex uncertainty a standard output of in-the-wild mesh recovery.

%% file: statements.tex
\subsection*{AI use statement}

In this work, we used generative AI tools for minor edits in writing. This includes finding ways to convey our thoughts in a compact fashion. GenAI tools were also used for grammatical checks or minor corrections. Bulk of the paper was written by humans, for humans. No AI was used in writing this statement. We additionally used GenAI tools for code implementation and documentation. However, there was always a human-in-the-loop to verify code correctness. Finally, GenAI was used to double-check the math in this paper. However, the authors developed the idea and formulated the derivation. We take responsibility for the final content of this work,
including text, claims or artifacts produced with the aid of generative AI.

\subsection*{Reproducibility statement}

We provide significant implementation details: the pseudo-code for \methodnameshort, and the variance head when using \methodnameshort for in-the-wild human mesh recovery. We also use the open-weight SAM 3D Body model as well as publicly available datasets. Barring our study on downstream decision making, all other experiments can be reproduced with publicly available data, models, and the implementation details provided in the paper.

%% file: appendix.tex
\section{Derivation Details}
\label{app:derivation}
We provide a detailed, annotated version of the derivation for convenience. We also provide code in the accompanying repository for 3DMM specific modifications, and present our general analysis here.

\paragraph{Assumptions and approximations.}
CLUE uses the law of total variance and first-order error propagation specialised to the skinning equation. We make several assumptions that allow this approximation to be real-time yet accurate in comparison to large-scale sampling and jacobian based error propagation.

\begin{enumerate}
\item \textit{Independence and diagonal covariance.} Following literature, $\beta$, $\psi$ and $\theta$ are mutually independent with diagonal covariances, and the perturbations of different joints are independent with zero mean. The pose uncertainty of joint $a$ is a Gaussian on its axis-angle coordinates $\omega_a$ and is expressed in the axes of the parent of $a$. The predicted per-axis variances $\sigma^2_{\theta_a}$ and its effect on the rotation is a small rotation composed with the mean rotation. We keep only per-vertex variances and not covariances for simplicity.

\item \textit{First order in the pose.} We drop terms of second or higher order in the pose perturbations. In particular $\mathbb{E}_\theta[\mathbf{G}_j]=\bar{\mathbf{G}}_j+O(\sigma_\theta^2)$, and we drop the $O(\sigma_\theta^2\sigma_X^2)$ cross term. This simplicity allows us to have a lower computational cost in relative to full error propagation.

\item \textit{Joint locations at the mean shape.}
When analyzing the shape and expression variance, the joint locations $\mathbf{J}$ and hence $\bar{\mathbf{t}}_j$ are held at the mean.

\item \textit{No pose correctives.} We don't model pose correctives, but show how they can be added if needed in this derivation.

\end{enumerate}

\subsection{Preliminary: 3DMM Formulation}
 
\textit{Rest-pose vertex.} Vertex $i$ in the rest pose is the template vertex with additive linear offsets of shape and expression,
\[
    X_i = \bar{v}_i + \vec{\beta}\, B_{S,i} + \vec{\psi}\, B_{E,i} \,.
\]
Here, $\beta \in \mathbb{R^{|\beta|}}$ and $B_{S, i} \in \mathbb{R}^{|\beta| \times 3}$ and similar for the expression component. 

\textit{Kinematic chain.} Each joint $k$ has a parent $p(k)$, a rest-pose location $\mathbf{J}_k$, an offset $\mathbf{j}_k = \mathbf{J}_k - \mathbf{J}_{p(k)}$ from its parent, and a local rotation $\exp(\omega_k)$ given by the pose $\theta = \{\omega_k\}_{k \in \mathcal{J}}$, where $\mathcal{J}$ is the set of joints. Let $A(j)$ be the ordered set of joints on the path from the root to $j$, including $j$. The global transformation of joint $j$ is the chain product
\[
    G_j(\theta, \mathbf{J}) = \prod_{k \in A(j)} \begin{bmatrix} \exp(\omega_k) & \mathbf{j}_k \\ 0 & 1 \end{bmatrix}
    =: \begin{bmatrix} \mathbf{R}_j & \mathbf{t}_j \\ 0 & 1 \end{bmatrix} ,
\]
where $\mathbf{R}_j$ is the global rotation of joint $j$ and $\mathbf{t}_j$ its location in the posed body. The chain can be represented as a recursive relation between a joint and its parent, $\mathbf{R}_j = \mathbf{R}_{p(j)} \exp(\omega_j)$ and $\mathbf{t}_j = \mathbf{R}_{p(j)}\, \mathbf{j}_j + \mathbf{t}_{p(j)}$.
 
We use two properties. First, $\omega_a$ appears in $G_j$ if and only if $a \in A(j)$; we write $\mathcal{D}(a) = \{ j : a \in A(j) \}$ for the joints at or below $a$. Second, each factor is a translation followed by a rotation,
\[
    \begin{bmatrix} \exp(\omega_k) & \mathbf{j}_k \\ 0 & 1 \end{bmatrix} = T(\mathbf{j}_k)\, \mathrm{Rot}\big(\exp(\omega_k)\big) \,,
    \qquad
    T(\mathbf{u}) = \begin{bmatrix} I & \mathbf{u} \\ 0 & 1 \end{bmatrix}, \quad
    \mathrm{Rot}(\mathbf{R}) = \begin{bmatrix} \mathbf{R} & 0 \\ 0 & 1 \end{bmatrix} .
\]
 
\textit{Pivot removal.} SMPL applies $G_j$ after removing the rest-pose transformation, $\mathbf{G}_j = G_j(\theta, \mathbf{J})\, G_j(\theta^*, \mathbf{J})^{-1}$. In the rest pose all rotations are the identity, so $G_j(\theta^*, \mathbf{J}) = T(\mathbf{J}_j)$ and
\[
    \mathbf{G}_j\, v = \mathbf{R}_j\, (v - \mathbf{J}_j) + \mathbf{t}_j \,, \qquad \text{in particular} \qquad \mathbf{G}_j\, \mathbf{J}_j = \mathbf{t}_j \,.
\]
 
\textit{Skinning.} The posed vertex is a weighted average of the vertex transformed by each joint,
\begin{equation}
    v'_i = \sum_{j \in \mathcal{J}} w_{i,j}\, \mathbf{G}_j\, X_i \,,
    \label{eq:app_skinning}
\end{equation}
with $w_{i,j}$ the skinning weights, which are non-zero only for a few joints near vertex $i$.
 
\subsection{Additive Variance.}
We return to the formulation of the rest pose vertex. The shape and expression parameters have variances $\sigma^2_\beta$ and $\sigma^2_\psi$. Since $X_i$ is linear in these parameters and using $\text{Var}[AZ] = A\Sigma_Z A^T$, the covariance of the rest-pose vertex is
\begin{equation}
    \Sigma_{X_i} = \text{Var}[X_i] = B_{S,i}^T\, \mathrm{diag}(\sigma^2_\beta)\, B_{S,i} + B_{E,i}^T\, \mathrm{diag}(\sigma^2_\psi)\, B_{E,i} \;.
    \label{eq:app_var_blendshapes}
\end{equation}
We denote the entries of $\Sigma_{X_i}$ by $\Sigma^{(kl)}_{X_i}$, $k, l \in \{1,2,3\}$.

\subsection{Law of Total Variance.}
We have two independent sources of randomness in the posed vertex $v'_i$: namely the shape and expression parameters, through $X_i$, and the pose $\theta$, through the transformations $\mathbf{G}_j$. The law of total variance separates them,
\begin{equation}
    \text{Var}(v'_i) = \underbrace{\mathbb{E}_X\big[ \text{Var}_\theta(v'_i \mid X_i) \big]}_{\text{rotation variance}} + \underbrace{\text{Var}_X\big( \mathbb{E}_\theta[v'_i \mid X_i] \big)}_{\text{shape and expression variance}} \,.
    \label{eq:app_ltv}
\end{equation}
The first term asks that for a fixed rest-pose vertex, how much does the vertex move when the pose varies? The second asks whether for the expected pose, how much does the vertex move when the shape and expression vary? We derive the two terms in turn.
 
\subsection{Rotation Variance.}
The intuition is to fix the rest-pose vertex $X_i$ and let only the pose vary.
 
\subsubsection{Step 1: Effect of one joint's perturbation \textit{(joint $a$)} on one transformation ($j$)}
 
Let $a \in A(j)$. Splitting the chain product of $G_j$ at joint $a$ and using the translate-then-rotate form of the $a^{th}$ factor,
\[
    G_j = \underbrace{G_{p(a)}\, T(\mathbf{j}_a)}_{H_a}\; \mathrm{Rot}\big(\exp(\omega_a)\big) \prod_{k \in A(j),\; k \text{ below } a} \begin{bmatrix} \exp(\omega_k) & \mathbf{j}_k \\ 0 & 1 \end{bmatrix} ,
    \qquad
    H_a = \begin{bmatrix} \mathbf{R}_{p(a)} & \mathbf{t}_a \\ 0 & 1 \end{bmatrix} ,
\]
where the block form of $H_a$ follows from the recursion $\mathbf{t}_a = \mathbf{R}_{p(a)}\, \mathbf{j}_a + \mathbf{t}_{p(a)}$. $H_a$ is the rigid transformation that places the frame of joint $a$ in the world: rotation $\mathbf{R}_{p(a)}$, origin at the posed joint location $\mathbf{t}_a$.
 
Now let us perturb the rotation of joint $a$. We have a Gaussian on the axis-angle coordinates of the joint, $\omega_a + \varepsilon_a$ with $\varepsilon_a \sim \mathcal{N}(0, \mathrm{diag}(\sigma^2_{\theta_a}))$, in the axes of the parent of $a$. The perturbed rotation is the unperturbed one composed with a small rotation,
\begin{gather*}
    \exp(\omega_a + \varepsilon_a) = (I + [\delta_a]_\times)\, \exp(\omega_a) \,, \qquad
    \delta_a = J_l(\omega_a)\, \varepsilon_a \,, \\
    J_l(\omega) = I + \frac{1 - \cos\vartheta}{\vartheta^2}\, [\omega]_\times + \frac{\vartheta - \sin\vartheta}{\vartheta^3}\, [\omega]_\times^2 \,, \qquad \vartheta = \|\omega\| \,,
\end{gather*}

where $J_l$ is the left Jacobian of $SO(3)$, which maps an increment of the axis-angle coordinates to the equivalent small rotation. Substituting it in $G_j$ we have,
\[
    G_j^{\text{pert}} = \underbrace{H_a\, \mathrm{Rot}\big(I + [\delta_a]_\times\big)\, H_a^{-1}}_{\text{extra factor}}\; G_j \,:
\]
the perturbed transformation is the unperturbed one followed by an extra factor. To see what this factor does, follow a point $x$ through its three matrices, from right to left.
\begin{enumerate}
    \item[(i)] $H_a^{-1}$ expresses $x$ in the frame of joint $a$. The inverse of a rigid transformation subtracts the translation and undoes the rotation,
    \[
        y = H_a^{-1} x = \mathbf{R}_{p(a)}^T\, (x - \mathbf{t}_a) \,.
    \]
    \item[(ii)] $\mathrm{Rot}(I + [\delta_a]_\times)$ applies the small rotation in that frame,
    \[
        y' = (I + [\delta_a]_\times)\, y = y + \delta_a \times y \,.
    \]
    \item[(iii)] $H_a$ brings the result back to world coordinates,
    \[
        x' = H_a\, y' = \mathbf{R}_{p(a)}\, y' + \mathbf{t}_a
           = \underbrace{\mathbf{R}_{p(a)}\, y + \mathbf{t}_a}_{=\; x} \;+\; \mathbf{R}_{p(a)}\, (\delta_a \times y) \,.
    \]
\end{enumerate}
Using the property, $\mathbf{R}\,(\mathbf{u} \times \mathbf{v}) = (\mathbf{R}\mathbf{u}) \times (\mathbf{R}\mathbf{v})$ and (i), the last term equals $(\mathbf{R}_{p(a)} \delta_a) \times (\mathbf{R}_{p(a)}\, y) = (\mathbf{R}_{p(a)} \delta_a) \times (x - \mathbf{t}_a)$. Writing $\delta\omega_a = \mathbf{R}_{p(a)}\, \delta_a$ for the perturbation expressed in world coordinates, the extra factor acts as
\[
    x \;\longmapsto\; x + \delta\omega_a \times (x - \mathbf{t}_a) \,.
\]
Intuitively, perturbing the rotation of joint $a$ rotates every point carried by the chain below $a$ by the small angle $\delta\omega_a$ about the posed location $\mathbf{t}_a$ of joint $a$.
 
\textit{From the joint transformation to the vertex.} We write $F_a = H_a\, \mathrm{Rot}\big(I + [\delta_a]_\times\big)\, H_a^{-1}$ for the extra factor, so that $G_j^{\text{pert}} = F_a\, G_j$ and $F_a\, x = x + \delta\omega_a \times (x - \mathbf{t}_a)$ for any point $x$. To accurately locate the vertex, we need to adjust for the pivot. However, pivot removal for $G_j^{\text{pert}}$ and $G_j$ is the same since this step does not depend on the pose, perturbed or not. Therefore, $\mathbf{G}_j^{\text{pert}} = F_a\, \mathbf{G}_j$. Hence, $F_a$ acts on the posed vertex $\mathbf{G}_j X_i$ like on any other point. For every joint $j$ at or below $a$,
\[
    \mathbf{G}_j^{\text{pert}} X_i = F_a\,(\mathbf{G}_j X_i) = \mathbf{G}_j X_i + \delta\omega_a \times \big( \mathbf{G}_j X_i - \mathbf{t}_a \big) \,, \qquad j \in \mathcal{D}(a) \,.
\]
For $j \notin \mathcal{D}(a)$, $\omega_a$ does not appear in the chain of $G_j$, so there is no factor $F_a$ and $\mathbf{G}_j X_i$ is unchanged.
 
\subsubsection{Step 2: Effect of perturbing all joints}
 
Let every joint be perturbed. Each $a \in A(j)$ contributes an extra perturbation. Expanding the product and dropping products of two or more small terms, the displacements of Step 1 add up to,
 
\[
    \mathbf{G}_j^{\text{pert}}\, X_i \;\approx\; \bar{\mathbf{G}}_j X_i + \sum_{a \in A(j)} \delta\omega_a \times \big( \bar{\mathbf{G}}_j X_i - \bar{\mathbf{t}}_a \big) \,,
\]
where a bar denotes evaluation at the mean pose $\mu_\theta$. Substituting into the skinning equation (Eq.~\ref{eq:app_skinning}),
\[
    v'_i \;\approx\; \sum_{j \in \mathcal{J}} w_{i,j}\, \bar{\mathbf{G}}_j X_i
    \;+\; \sum_{j \in \mathcal{J}} \sum_{a \in A(j)} w_{i,j}\, \delta\omega_a \times \big( \bar{\mathbf{G}}_j X_i - \bar{\mathbf{t}}_a \big) \,.
\]
The double sum runs over the pairs $(j, a)$ with $a \in A(j)$, which is the same set of pairs as $j \in \mathcal{D}(a)$. Exchanging the order of summation therefore gives $\sum_{j \in \mathcal{J}} \sum_{a \in A(j)} = \sum_{a \in \mathcal{J}} \sum_{j \in \mathcal{D}(a)}$, and since $\delta\omega_a$ does not depend on $j$ it factors out of the inner sum
\[
    v'_i \;\approx\; \sum_{j \in \mathcal{J}} w_{i,j}\, \bar{\mathbf{G}}_j X_i
    \;+\; \sum_{a \in \mathcal{J}} \delta\omega_a \times \boldsymbol{\ell}_{i,a} \,,
    \qquad
    \boldsymbol{\ell}_{i,a} = \sum_{j \in \mathcal{D}(a)} w_{i,j}\, \big( \bar{\mathbf{G}}_j X_i - \bar{\mathbf{t}}_a \big) \,.
\]
For each joint $a$ it collects exactly those skinning terms whose joint lies at or below $a$, the only terms that $\omega_a$ affects. Since $\bar{\mathbf{t}}_a$ does not depend on $j$,
\[
    \boldsymbol{\ell}_{i,a} = \sum_{j \in \mathcal{D}(a)} w_{i,j}\, \bar{\mathbf{G}}_j X_i \;-\; W_{i,a}\, \bar{\mathbf{t}}_a \,,
    \qquad
    W_{i,a} = \sum_{j \in \mathcal{D}(a)} w_{i,j} \,,
\]
with $W_{i,a}$ the total skinning weight of vertex $i$ on the joints at or below $a$. We call $\boldsymbol{\ell}_{i,a}$ the lever arm of vertex $i$ about joint $a$. It vanishes when no joint in $\mathcal{D}(a)$ skins vertex $i$, so only the joints on the chains above vertex $i$ contribute.
 
\subsubsection{Step 3: variance} The first sum in Step 2 is deterministic whereas the second is the random. Writing $\delta\omega_a \times \boldsymbol{\ell}_{i,a} = -\boldsymbol{\ell}_{i,a} \times \delta\omega_a = -[\boldsymbol{\ell}_{i,a}]_\times\, \delta\omega_a$ and assuming the perturbations of different joints to be independent with zero mean,
\begin{gather*}
    \text{Var}_\theta(v'_i \mid X_i) = \sum_{a \in \mathcal{J}} [\boldsymbol{\ell}_{i,a}]_\times\, \Sigma_{\theta_a}\, [\boldsymbol{\ell}_{i,a}]_\times^T \,, \\
    \Sigma_{\theta_a} = \mathrm{Cov}(\delta\omega_a) = \bar{\mathbf{R}}_{p(a)}\,\mathrm{Var}(\delta_a)\,\bar{\mathbf{R}}_{p(a)}^{T} \,, \\
    \mathrm{Var}(\delta_a) = J_l(\bar{\omega}_a)\, \mathrm{diag}\big(\sigma^2_{\theta_a,x},\, \sigma^2_{\theta_a,y},\, \sigma^2_{\theta_a,z}\big)\, J_l(\bar{\omega}_a)^T \,.
\end{gather*}
Here, $\sigma^2_{\theta_a,\cdot}$ are the predicted per-axis variances of the axis-angle coordinates of joint $a$, in the axes of its parent, $J_l(\bar{\omega}_a)$ is the left Jacobian of Step 1 evaluated at the mean pose, and $\bar{\mathbf{R}}_{p(a)}$ is the mean global rotation of the parent. $\Sigma_{\theta_a}$ is the covariance in world axes, $\Sigma^{(pq)}_{\theta_a}$, $p, q \in \{x, y, z\}$. The outer expectation over $X_i$ in Eq.~\ref{eq:app_ltv} is taken to first order by evaluating the lever arm at the mean rest-pose vertex $\mu_{X_i}$,
\[
    \bar{\boldsymbol{\ell}}_{i,a} = \sum_{j \in \mathcal{D}(a)} w_{i,j}\, \bar{\mathbf{G}}_j\, \mu_{X_i} - W_{i,a}\, \bar{\mathbf{t}}_a = \big( \ell^{(x)}_{i,a},\, \ell^{(y)}_{i,a},\, \ell^{(z)}_{i,a} \big)^T .
\]
Writing $[\bar{\boldsymbol{\ell}}_{i,a}]_\times$ as the cross-product matrix of this vector, the rotation variance is the diagonal of $[\bar{\boldsymbol{\ell}}_{i,a}]_\times \Sigma_{\theta_a} [\bar{\boldsymbol{\ell}}_{i,a}]_\times^T$. This gives us
\begin{equation}
    \mathbb{E}_X\big[ \text{Var}_\theta(v'_i \mid X_i) \big] \approx \sum_{a \in \mathcal{J}}
\begin{bmatrix}
\Sigma^{(yy)}_{\theta_a} \big(\ell^{(z)}_{i,a}\big)^2 + \Sigma^{(zz)}_{\theta_a} \big(\ell^{(y)}_{i,a}\big)^2 - 2\, \Sigma^{(yz)}_{\theta_a}\, \ell^{(y)}_{i,a}\, \ell^{(z)}_{i,a} \\
\Sigma^{(xx)}_{\theta_a} \big(\ell^{(z)}_{i,a}\big)^2 + \Sigma^{(zz)}_{\theta_a} \big(\ell^{(x)}_{i,a}\big)^2 - 2\, \Sigma^{(xz)}_{\theta_a}\, \ell^{(x)}_{i,a}\, \ell^{(z)}_{i,a} \\
\Sigma^{(xx)}_{\theta_a} \big(\ell^{(y)}_{i,a}\big)^2 + \Sigma^{(yy)}_{\theta_a} \big(\ell^{(x)}_{i,a}\big)^2 - 2\, \Sigma^{(xy)}_{\theta_a}\, \ell^{(x)}_{i,a}\, \ell^{(y)}_{i,a}
\end{bmatrix} .
\label{eq:app_final_rotation}
\end{equation}

\subsection{Shape and Expression Variance.}
Now let the pose take its expected value and let only the shape and expression vary. To first order in the pose perturbation (A2), the expected transformation of each joint is the transformation at the mean pose, $\bar{\mathbf{G}}_j$. The expected vertex is then
\[
    \mathbb{E}_\theta[v'_i \mid X_i] = \Big( \sum_{j \in \mathcal{J}} w_{i,j}\, \bar{\mathbf{G}}_j \Big) X_i \,.
\]
We note that $X_i$ is random only in its first three components (the offsets). The joint locations $\mathbf{J}$ and the translations $\bar{\mathbf{t}}_j$ are evaluated at the mean shape and interact only with the homogeneous coordinate of 1 and do not contribute to the variance. What remains is the skinning-blended rotation
\[
    \mathbf{R}^{\mathrm{eff}}_i = \sum_{j \in \mathcal{J}} w_{i,j}\, \bar{\mathbf{R}}_j
\]
applied to the rest-pose offsets. Using $\text{Var}[AZ] = A\Sigma_Z A^T$ with the covariance $\Sigma_{X_i}$ of Eq.~\ref{eq:app_var_blendshapes}, the shape and expression variance is the diagonal of $\mathbf{R}^{\mathrm{eff}}_i\, \Sigma_{X_i}\, (\mathbf{R}^{\mathrm{eff}}_i)^T$,
\begin{equation}
    \text{Var}_X\big( \mathbb{E}_\theta[v'_i \mid X_i] \big) =
\begin{bmatrix}
\sum_{k,l=1}^{3} \mathbf{R}^{\mathrm{eff}}_{i,1,k}\, \mathbf{R}^{\mathrm{eff}}_{i,1,l}\, \Sigma^{(kl)}_{X_i} \\
\sum_{k,l=1}^{3} \mathbf{R}^{\mathrm{eff}}_{i,2,k}\, \mathbf{R}^{\mathrm{eff}}_{i,2,l}\, \Sigma^{(kl)}_{X_i} \\
\sum_{k,l=1}^{3} \mathbf{R}^{\mathrm{eff}}_{i,3,k}\, \mathbf{R}^{\mathrm{eff}}_{i,3,l}\, \Sigma^{(kl)}_{X_i}
\end{bmatrix} .
\label{eq:app_final_blendshape}
\end{equation}
Intuitively, the rest-pose offset covariance of the vertex is rotated into the posed body by the vertex's own blended rotation.
 
Finally, substituting Eq.~\ref{eq:app_final_rotation} and Eq.~\ref{eq:app_final_blendshape} into Eq.~\ref{eq:app_ltv} gives the per-vertex variance
\begin{equation}
\begin{aligned}
    \text{Var}(v'_i) \approx{}&
\sum_{a \in \mathcal{J}}
\begin{bmatrix}
\Sigma^{(yy)}_{\theta_a} \big(\ell^{(z)}_{i,a}\big)^2 + \Sigma^{(zz)}_{\theta_a} \big(\ell^{(y)}_{i,a}\big)^2 - 2\, \Sigma^{(yz)}_{\theta_a}\, \ell^{(y)}_{i,a}\, \ell^{(z)}_{i,a} \\
\Sigma^{(xx)}_{\theta_a} \big(\ell^{(z)}_{i,a}\big)^2 + \Sigma^{(zz)}_{\theta_a} \big(\ell^{(x)}_{i,a}\big)^2 - 2\, \Sigma^{(xz)}_{\theta_a}\, \ell^{(x)}_{i,a}\, \ell^{(z)}_{i,a} \\
\Sigma^{(xx)}_{\theta_a} \big(\ell^{(y)}_{i,a}\big)^2 + \Sigma^{(yy)}_{\theta_a} \big(\ell^{(x)}_{i,a}\big)^2 - 2\, \Sigma^{(xy)}_{\theta_a}\, \ell^{(x)}_{i,a}\, \ell^{(y)}_{i,a}
\end{bmatrix} \\
&+
\begin{bmatrix}
\sum_{k,l=1}^{3} \mathbf{R}^{\mathrm{eff}}_{i,1,k}\, \mathbf{R}^{\mathrm{eff}}_{i,1,l}\, \Sigma^{(kl)}_{X_i} \\
\sum_{k,l=1}^{3} \mathbf{R}^{\mathrm{eff}}_{i,2,k}\, \mathbf{R}^{\mathrm{eff}}_{i,2,l}\, \Sigma^{(kl)}_{X_i} \\
\sum_{k,l=1}^{3} \mathbf{R}^{\mathrm{eff}}_{i,3,k}\, \mathbf{R}^{\mathrm{eff}}_{i,3,l}\, \Sigma^{(kl)}_{X_i}
\end{bmatrix} .
\end{aligned}
\label{eq:app_final}
\end{equation}

\subsection{Pose Correctives}
\label{app:pose_correctives}

We use STAR as the 3DMM to analyse pose correctives. Following STAR's notation convention, $P_j$ is the per-joint pose corrective function which regresses corrective offsets given the joint and its direct neighbors’ quaternion values. 
\[P_j(\boldsymbol{q}_{ne(j)}, \beta_2; \mathcal{K}_j)
  = \mathcal{K}_j \left( (\boldsymbol{q}_{ne(j)} - \boldsymbol{q}^{*}_{ne(j)})^{T} \,|\, \beta_2 \right)^{T}
\]
Here, the second principal component $\beta_2$ as well as $\boldsymbol{q}_{ne(j)}$ are stochastic, while $\boldsymbol{q}_{ne(j)}^{*}$ and $\mathcal{K}_j$ are deterministic. We note that the variance of this equation follows the familiar $\text{Var}\big[ AZ \big] = A \Sigma_Z A^T$ format where $A$ is a deterministic matrix and $Z$ is random.

\begin{equation}
B_P^j(\boldsymbol{q}_{ne(j)}; \mathbf{A}_j, \mathcal{K}_j)
  = \phi(\mathbf{A}_j) \circ P_j(\boldsymbol{q}_{ne(j)}, \beta_2; \mathcal{K}_j)
\end{equation}

Since the activation weights $\mathbf{A}_j$ are fixed, $\phi(\mathbf{A}_j)$ is a deterministic vector. The variance of a hadamard product is $\text{Var}\big[ A \odot Z \big]_i = a^2_i \sigma^2_{Z_i}$. This allows us to compute the variance from pose correctives.

\section{Training Details}

\label{app:training}

\textbf{Training.} All methods share the same training recipe and differ only in the probabilistic modeling. The DINOv3 ViT-H/16+ image encoder is initialised and kept frozen. All other modules of the SAM 3D Body architecture are randomly initialized and trained from scratch. We optimise with AdamW (learning rate $10^{-3}$, $\beta=(0.9,0.999)$, weight decay $0.05$), a linear warm-up over the first 1k steps followed by cosine decay to 5\% of the peak learning rate, and gradient clipping at norm $1.0$. Each step processes 32 images per GPU on 4 GPUs (128 images per step); we train for 30k steps. To train SAM 3D Body (\methodnameshort), we freeze SAM 3D Body and only keep the variance head learnable. We train this head for 10k steps.

\textbf{Losses.} All methods share the same deterministic terms: an L2 loss on the 2D keypoints using the predicted perspective camera, an L2 loss on the 3D joints, and a joint-limit penalty on body joint angles. The methods differ only on how the MHR parameters are supervised. Deterministic training uses L2 loss on all MHR parameters. Parameter-space NLL replaces these deterministic losses on the MHR parameters. Normalizing flow applies NLL and RealNVP flow NLL on the joint parameters, keeping shape loss deterministic. Matrix-Fisher uses a matrix-Fisher NLL on each supervised joint rotation, with concentrations predicted hierarchically along the kinematic chain, and the Gaussian NLL for lengths and shape. \methodnameshort uses the same losses as parameter-space NLL, but replaces L2 loss over the joints with NLL. All loss terms have unit weight except the joint-limit penalty ($0.01$).

\textbf{Metrics.} We follow standard practices and report (1) joint based metrics such as MPJPE and PA-MPJPE, and (2) vertex based metrics such as PVE and PA-PVE. When the prediction and (pseudo) ground truth are in the same space (MHR space), we additionally report variance based metrics such as vertex-NLL, spearman and pearson correlation AUSE. Since baselines measure uncertainty in the parameter space, we draw 100 samples per prediction to compute the mesh space variance.

\textbf{Datasets.} For training, we use the Segment Anything-1B (SA-1B) dataset which consists of 1.2 million+ training images \cite{kirillov2023segany}. Additionally, we use MPII and COCO with MHR annotations for training. For our evaluation, we use 3DPW \cite{vonMarcard2018} as our primary dataset containing ground truth motion capture data in SMPL format. Since our predictions are in MHR space and the ground truth is in SMPL, we use on-the-fly optimzation to map MHR to SMPL as done in \cite{yang2026sam, ferguson2025mhr}. Additionally, we also use MPII \cite{andriluka14cvpr}, Human3.6M \cite{h36m_pami}, COCO \cite{lin2014microsoft}, MPI-INF-3D \citep{mono-3dhp2017}, PoseTrack \cite{Andriluka_2018_CVPR}, CrowdPose \cite{DBLP:conf/cvpr/LiWZMFL19} and OCHuman \cite{zhang2019pose2seg}. Unlike 3DPW, these datasets do not have 3DMM annotations, and hence we use SAM 3D Body as the pseudo ground truth.

\section{Pseudo-Code}

\label{app:pseudocode}

We present the algorithm for an efficient implementation of \methodnamelong in Algorithm~\ref{alg:CLUE}. 

\input{algorithm}

\section{Full Results}

We gather our results and visualizations in this section.

\input{tables/pGT}

\begin{table}
\centering
\footnotesize
\setlength{\tabcolsep}{3pt}
\renewcommand{\arraystretch}{1.05}
\caption{\textbf{Compute cost across body models.} Time (ms), memory (MB) and GFLOPs of one variance evaluation. Median of 10 trials after 3 warm-ups on one CPU (Intel i7-8700).}
\label{tab:compute-cost}
\begin{tabular}{l p{6pt} rrr p{6pt} rrr p{6pt} rrr}
\toprule
 & & \multicolumn{3}{c}{\textbf{FLAME}} & & \multicolumn{3}{c}{\textbf{SMPL-X}} & & \multicolumn{3}{c}{\textbf{MHR}} \\
\cmidrule(lr){3-5} \cmidrule(lr){7-9} \cmidrule(lr){11-13}
\textbf{Method} & & ms & MB & GFLOPs & & ms & MB & GFLOPs & & ms & MB & GFLOPs \\
\midrule
Sampling $N{=}100$   & & 141.6  & 14.2  & 1.43  & & 368.5  & 29.5  & 5.42  & & 681.7  & 44.5   & 1.71 \\
Sampling $N{=}1000$  & & 1404.7 & 140.9 & 14.33 & & 3733.6 & 293.6 & 54.20 & & 6802.4 & 442.9  & 17.09 \\
Jacobian             & & 228.6  & 410.4 & 6.08  & & 177.0  & 383.6 & 10.23 & & 3272.2 & 1984.4 & 6.40 \\
\midrule
\rowcolor{gray!12}
\textbf{CLUE (ours)} & & \textbf{3.5} & \textbf{4.3} & \textbf{0.039} & & \textbf{4.0} & \textbf{9.5} & \textbf{0.015} & & \textbf{10.9} & \textbf{16.9} & \textbf{0.055} \\
\bottomrule
\end{tabular}
\end{table}

\begin{figure}
\centering
\vspace*{-10mm}
\begin{subfigure}{0.8\textwidth}
\includegraphics[width=\linewidth]{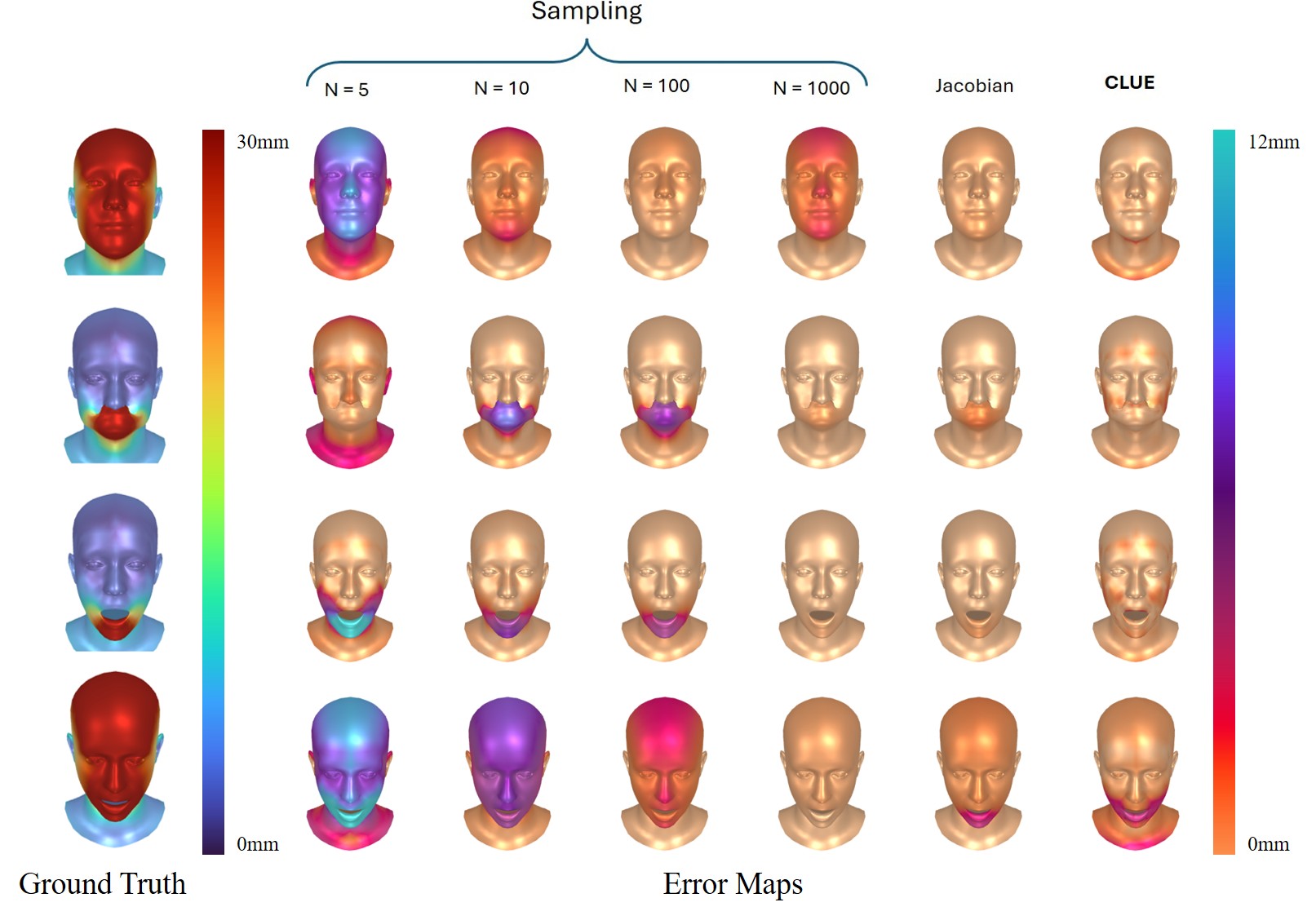}
\caption{FLAME}
\label{fig:flame_app}
\end{subfigure}

\begin{subfigure}{0.8\textwidth}
\includegraphics[width=\linewidth]{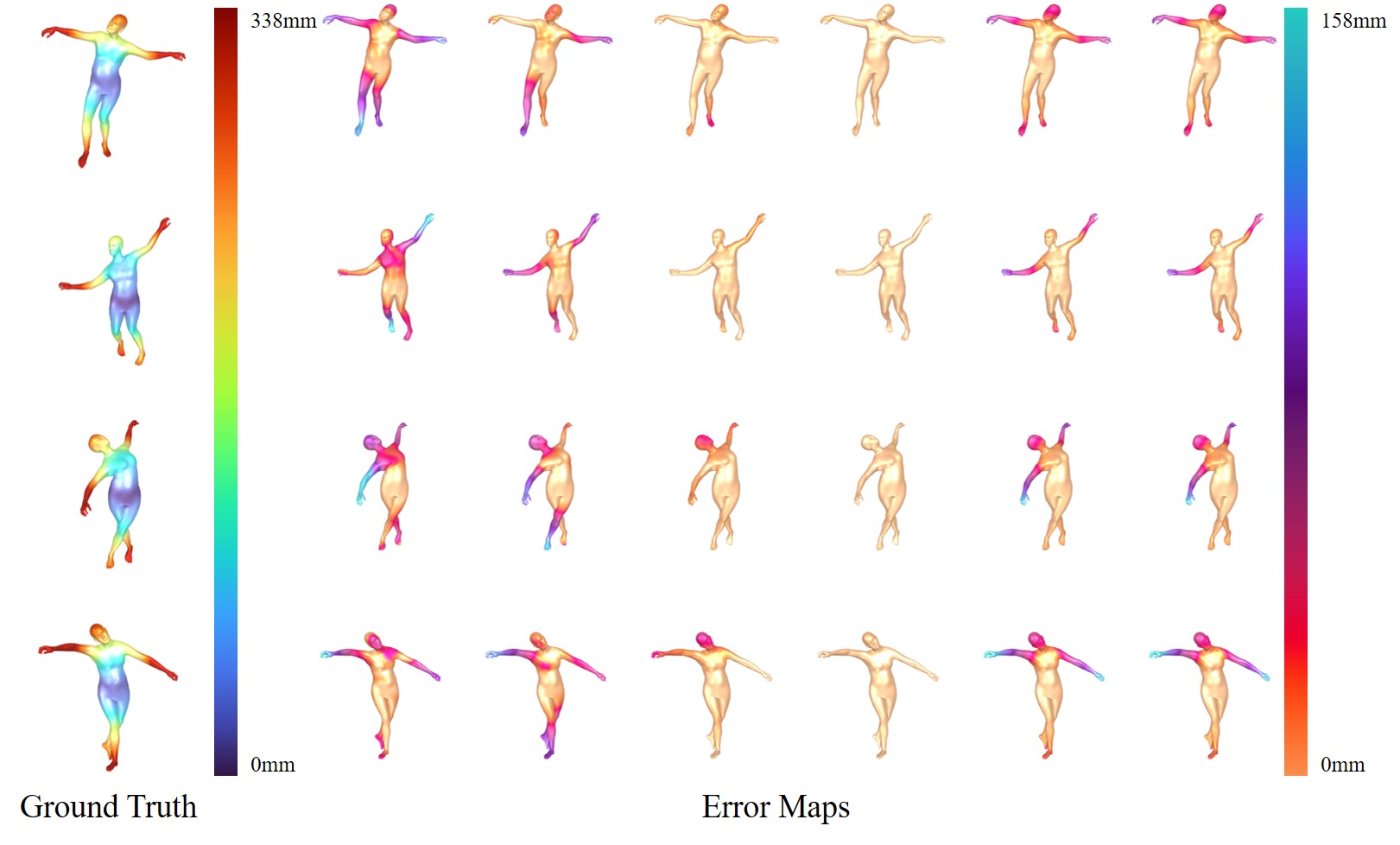}
\caption{SMPLX}
\label{fig:smplx_app}
\end{subfigure}

\begin{subfigure}{0.8\textwidth}
\includegraphics[width=\linewidth]{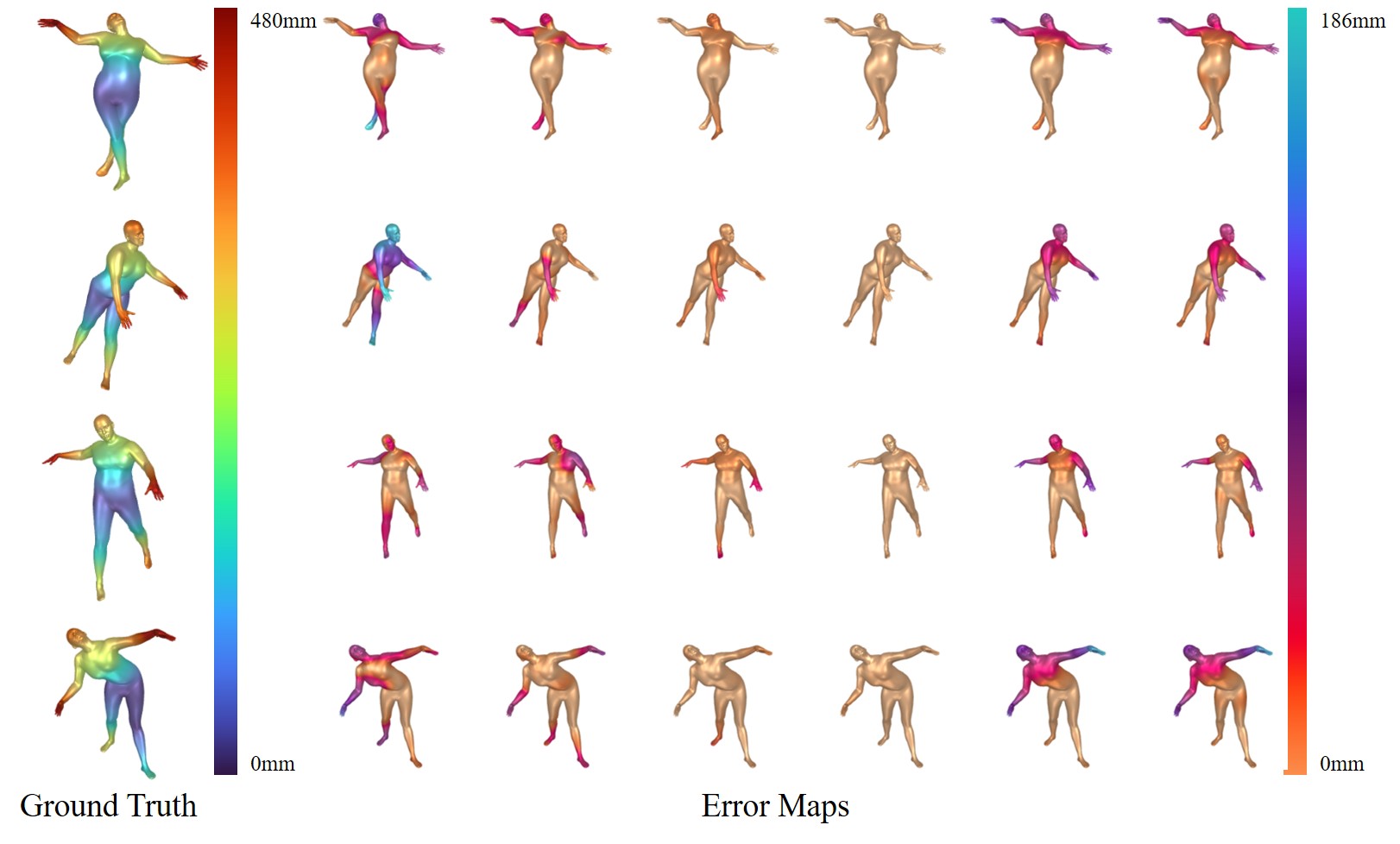}
\caption{MHR}
\label{fig:mhr_app}
\end{subfigure}
\vspace*{-3mm}
\caption{Complete error maps for FLAME, SMPLX, MHR. Both, Jacobian based linear error propagation and CLUE overestimate the variance due to first order assumptions. However when used in training, the learnt parameter variances compensate for them.}
\label{fig:error_maps_app}
\end{figure}

\begin{figure}
    \centering
    \includegraphics[width=\linewidth]{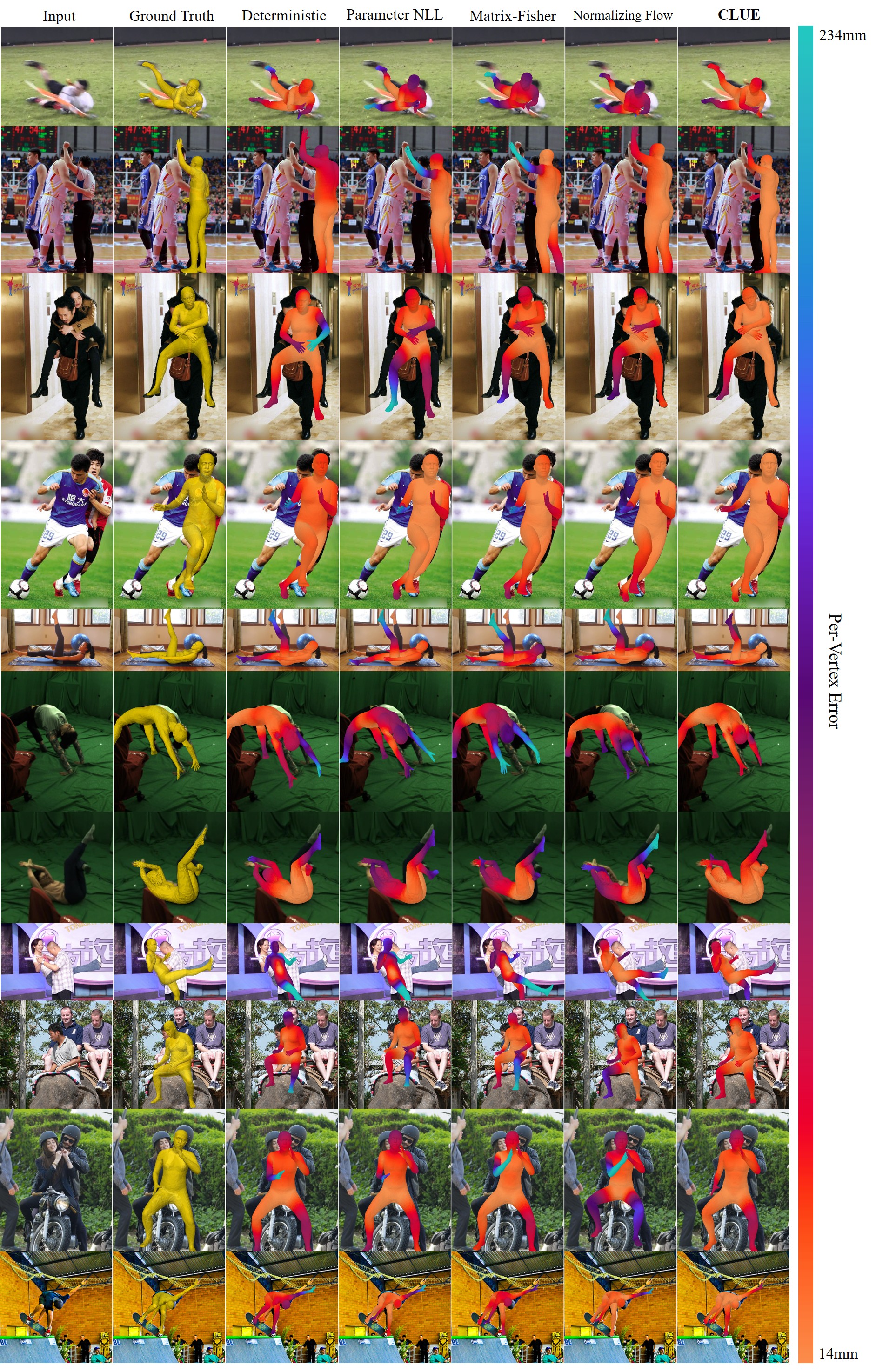}
    \caption{Error Maps for in-the-wild Human Mesh Recovery. We show that \methodnameshort has higher accuracy when dealing with occlusions or complex poses.}
    \label{fig:itw_errormaps_appendix}
\end{figure}

\begin{figure}
    \vspace*{-12mm}
    \centering
    \includegraphics[width=\linewidth]{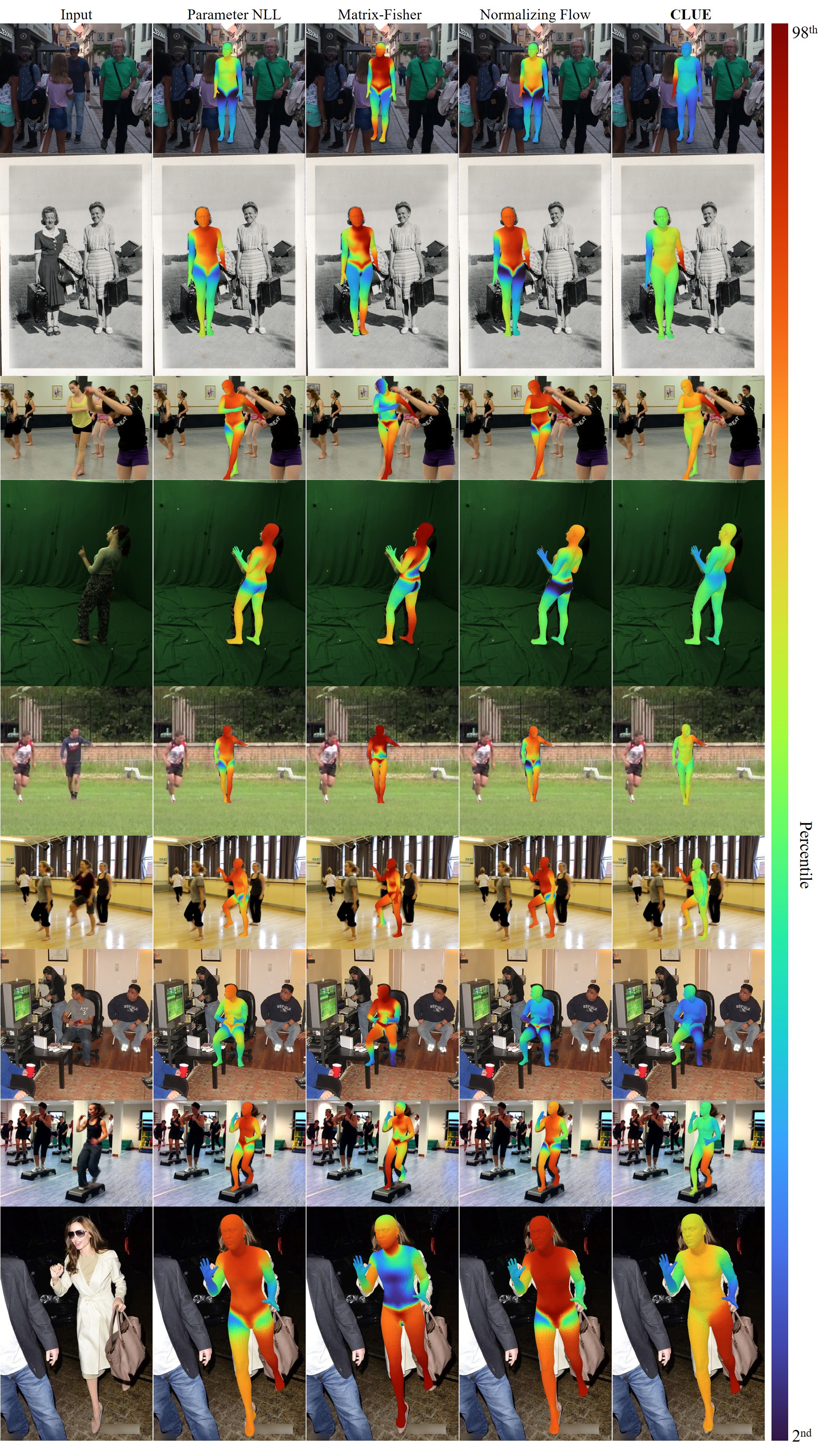}
    \caption{Uncertainty based data-mining.}
    \label{fig:datamining_appendix}
\end{figure}

\begin{figure}
    \vspace*{-12mm}
    \centering
    \includegraphics[width=0.9\linewidth]{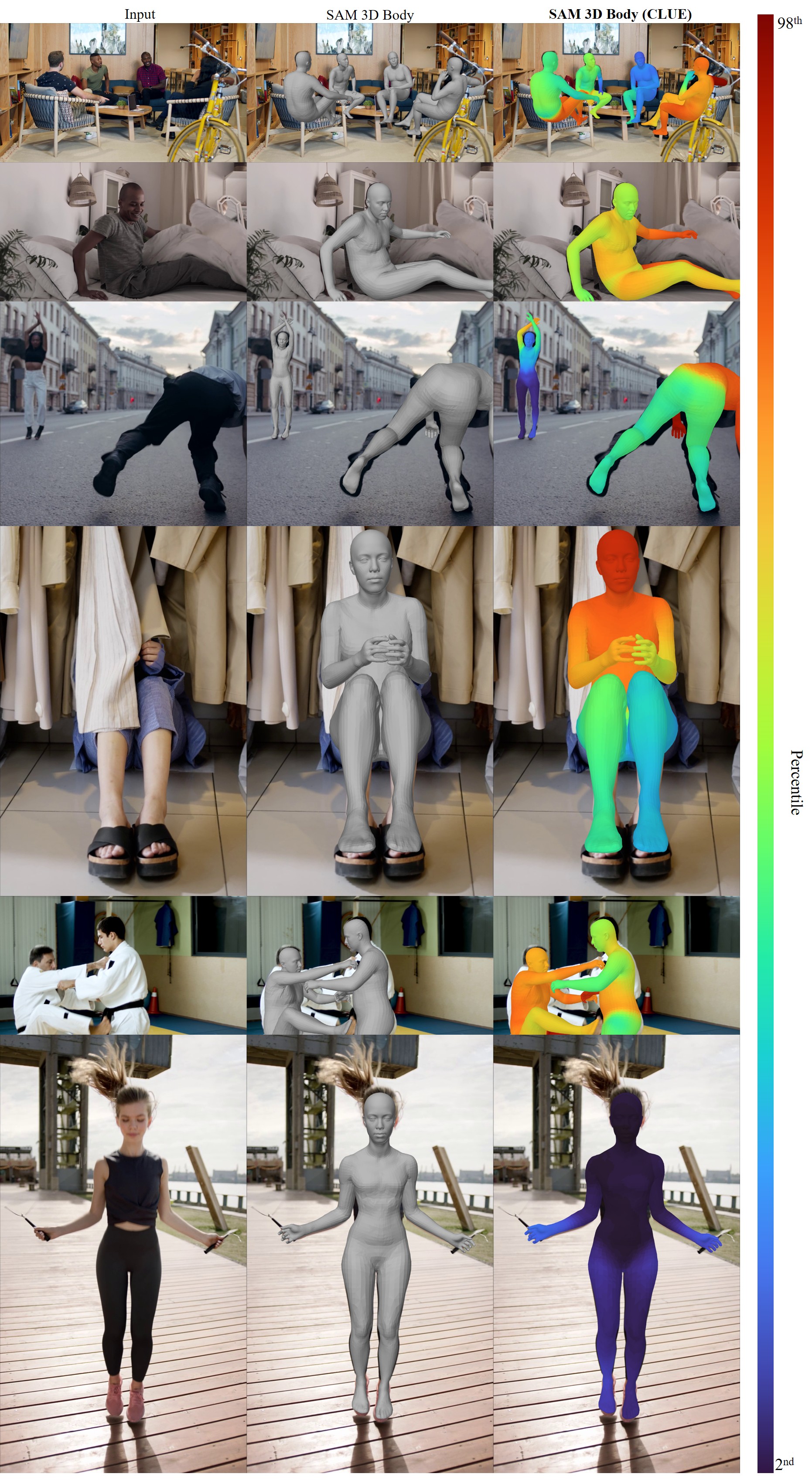}
    \caption{Augmenting SAM 3D Body with probabilistic reasoning using CLUE (Part 1).}
    \label{fig:sam3dbody_clue_1}
\end{figure}

\begin{figure}
    \vspace*{-12mm}
    \centering
    \includegraphics[width=0.95\linewidth]{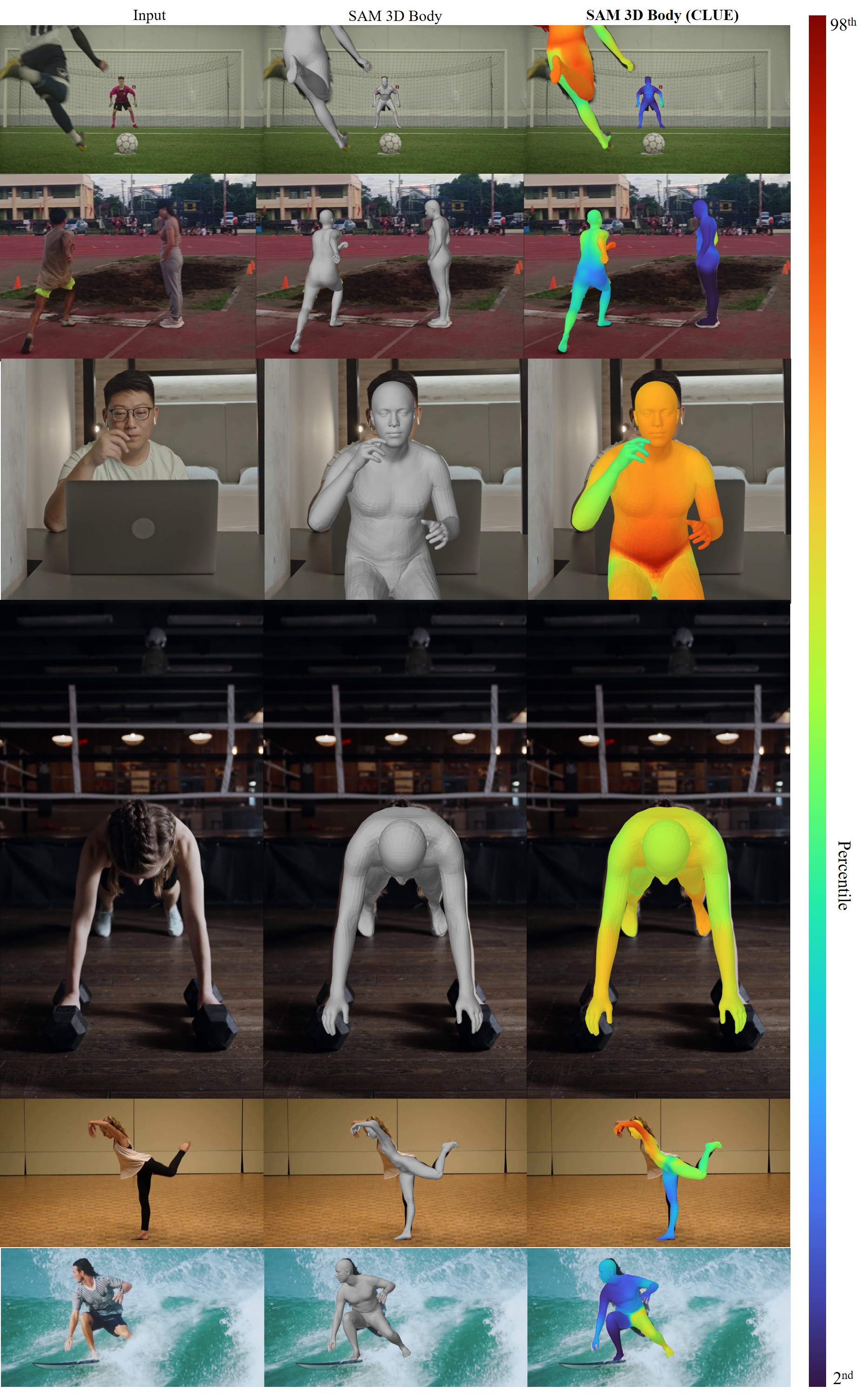}
    \caption{Augmenting SAM 3D Body with probabilistic reasoning using CLUE (Part 2).}
    \label{fig:sam3dbody_clue_2}
\end{figure}

%% file: algorithm.tex
\begin{algorithm}[t]
\caption{\methodnameshort}
\label{alg:CLUE}
\begin{algorithmic}[1]
\Require{%
    $\boldsymbol{\sigma}^2_{\beta} \in \mathbb{R}^{B \times C}$: shape and expression parameter variances \textit{concatenated};
    $\boldsymbol{\sigma}^2_\theta \in \mathbb{R}^{B \times J \times 3}$: pose variances, per axis-angle coordinate in the parent's axes;
    $\bar{v} \in \mathbb{R}^{V \times 3}$: template mesh;
    $\mathbf{B} \in \mathbb{R}^{V \times 3 \times C}$: blendshape basis (shape $\cup$ expression);
    $\mathcal{R} \in \mathbb{R}^{J \times V}$: joint regressor;
    $W \in \mathbb{R}^{V \times J}$: LBS weights;
    $\vec{\beta} \in \mathbb{R}^{B \times C}$: mean shape and expression coefficients concatenated;
    $\vec{\theta} \in \mathbb{R}^{B \times J \times 3}$: mean pose parameters (axis-angle);
    $\pi$: parent map
}
\Ensure{$\mathrm{Var}(v'_i) \in \mathbb{R}^{B \times V \times 3}$: per-vertex variance}
 
\Statex
\Statex \textit{--- Precompute once at initialisation to reduce runtime ---}
 
\State \textbf{Build ancestor matrix} $P \in \{0,1\}^{J \times J}$:
\For{$k = 0, \ldots, J{-}1$}
    \State $j \gets k$
    \While{$j \geq 0$} \Comment{No parent is represented as -1}
        \State $P_{k,j} \gets 1$;\quad $j \gets \pi(j)$
    \EndWhile
\EndFor
\Statex
\State $\mathbf{B}^{\otimes}_{v,k,l,m} \gets \mathbf{B}_{v,k,m}\,\mathbf{B}_{v,l,m}$
    \hfill \Comment{$(V \times 3 \times 3 \times C)$, outer product per vertex and blendshape}
\State $\widetilde{W} \gets W[:,\, :,\,\text{None}] \cdot P[\text{None},\,:,\,:]$
    \hfill \Comment{Fusing skinning weights and ancestor matrix}
 
\Statex
\Statex \textit{--- Per-forward pass ---}
 
\Statex \textbf{Step 1: Shape / expression covariance} \hfill \Comment{Eq.~\ref{eq:app_var_blendshapes}}
\State $(\Sigma_X)_{b,v,k,l} \gets \sum_m \sigma^2_\beta[b,m]\,\mathbf{B}^{\otimes}_{v,k,l,m}$ \Comment{$(B \times V \times 3 \times 3)$, offset covariance}
\Statex
\Statex \textbf{Step 2: Forward kinematics}
\State $\hat{v} \gets \bar{v} + \mathbf{B}\,\vec{\beta}$
    \hfill \Comment{$(B \times V \times 3)$, shaped mesh}
\State $\mathbf{J} \gets \mathcal{R}\,\hat{v}$
    \hfill \Comment{$(B \times J \times 3)$, joint locations}
\State $\{\mathbf{R}_j, \mathbf{t}_j\}_{j=1}^{J} \gets \mathrm{FK}(\vec{\theta}, \mathbf{J}, \pi)$
    \hfill \Comment{rotations and translations of the pivot-removed $\mathbf{G}_j$}
\State $\mathbf{T}_j \gets \mathbf{R}_j\,\mathbf{J}_j + \mathbf{t}_j \quad \forall j$
    \hfill \Comment{$(B \times J \times 3)$, posed joint centres $\mathbf{G}_j \mathbf{J}_j$ (pivots)}
 
\Statex
\Statex \textbf{Step 3: Vertex transformation}
\Statex \textit{Modelling effect of different joint transformations on the vertex}
\State $\tilde{v} \gets \mathbf{R}\,\hat{v} + \mathbf{t}$
    \hfill \Comment{\texttt{bjpq,bvq->bvjp};\quad $(B \times V \times J \times 3)$, $\mathbf{G}_j \hat{v}$}
\Statex \textit{Broadcast helps in parallelizing offset calculation w.r.t pivot}
\State $r \gets \tilde{v}[:,\,:,\,:,\,\text{None}, :] - \mathbf{T}[:,\,\text{None},\,\text{None},\,:,\,:]$
    \hfill \Comment{rotation around pivot;\quad $(B \times V \times K \times J \times 3)$}
\Statex \textit{Reducing this w.r.t fused skinning weights + ancestor matrix}
\State $\mathbf{r}_{b,v,j,3} \gets \displaystyle\sum_k \tilde{W} \cdot r$
    \hfill \Comment{\texttt{vkj,bvkjd->bvjd};\quad $(B \times V \times J \times 3)$, lever arm $\bar{\boldsymbol{\ell}}_{v,j}$ about joint $j$}
 
\Statex
\Statex \textbf{Step 4: Rotational variance} \hfill \Comment{Eq.~\ref{eq:app_final_rotation}}
\State $a_j,\,b_j,\,c_j \gets \mathbf{r}_{v,j}^{(x)},\; \mathbf{r}_{v,j}^{(y)},\; \mathbf{r}_{v,j}^{(z)}$
    \hfill \Comment{$(B \times V \times J)$ each}
\State $J_l(\vec{\theta}_j) \gets I + \frac{1 - \cos\vartheta_j}{\vartheta_j^2}\,[\vec{\theta}_j]_\times + \frac{\vartheta_j - \sin\vartheta_j}{\vartheta_j^3}\,[\vec{\theta}_j]_\times^2,\quad \vartheta_j = \|\vec{\theta}_j\|$
    \hfill \Comment{$(B \times J \times 3 \times 3)$, left Jacobian of $SO(3)$ at the mean pose}
\State $\Sigma_j \gets \mathbf{R}_{\pi(j)}\, J_l(\vec{\theta}_j)\, \mathrm{diag}\big(\boldsymbol{\sigma}^2_{\theta,j}\big)\, J_l(\vec{\theta}_j)^T\, \mathbf{R}_{\pi(j)}^T$
    \hfill \Comment{$(B \times J \times 3 \times 3)$, world axes, $\mathbf{R}_{\pi(\text{root})} = I$}
\State $\mathrm{Var}_\theta(v') \gets \displaystyle\sum_{j=1}^{J}
\begin{bmatrix}
    \Sigma^{(yy)}_j\,c_j^2 + \Sigma^{(zz)}_j\,b_j^2 - 2\,\Sigma^{(yz)}_j\,b_j c_j \\[4pt]
    \Sigma^{(xx)}_j\,c_j^2 + \Sigma^{(zz)}_j\,a_j^2 - 2\,\Sigma^{(xz)}_j\,a_j c_j \\[4pt]
    \Sigma^{(xx)}_j\,b_j^2 + \Sigma^{(yy)}_j\,a_j^2 - 2\,\Sigma^{(xy)}_j\,a_j b_j
\end{bmatrix}$
    \hfill \Comment{$(B \times V \times 3)$}
\Statex
\Statex \textbf{Step 5: Rotated shape / expression variance} \hfill \Comment{Eq.~\ref{eq:app_final_blendshape}}
\State $\mathbf{R}^{\mathrm{eff}}_{v} \gets \displaystyle\sum_{j=1}^{J} w_{v,j}\,\mathbf{R}_j$
    \hfill \Comment{$(B \times V \times 3 \times 3)$, LBS-blended rotation}
\State $\mathrm{Var}_\beta(v')_p \gets \displaystyle\sum_{k,l=1}^{3} \mathbf{R}^{\mathrm{eff}}_{v,p,k}\,\mathbf{R}^{\mathrm{eff}}_{v,p,l}\,(\Sigma_X)_{v,k,l} \quad \forall p \in \{1,2,3\}$
    \hfill \Comment{$(B \times V \times 3)$}
 
\Statex \textbf{Step 6: Total variance} \hfill \Comment{Eq.~\ref{eq:app_final}}
\State \Return $\mathrm{Var}(v') \gets \mathrm{Var}_\theta(v') + \mathrm{Var}_\beta(v')$
\end{algorithmic}
\end{algorithm}

%% file: tables/pGT.tex
\newcolumntype{G}{>{\columncolor{gray!12}}c}   % tinted "ours" columns

\begin{table}[t]
\caption{\textbf{Pseudo-ground-truth evaluation on five in-the-wild
datasets}: (a) joint accuracy, (b) surface accuracy (mm, $\downarrow$),
(c) uncertainty quality — per-vertex NLL ($\downarrow$) and the
Pearson correlation between predicted $\sigma$ and vertex error
($\uparrow$), and (d) the area under the sparsification error curve (AUSE, $\downarrow$; per person, ranking by the mean predicted vertex $\sigma$, normalised by each dataset's full-set error).}
\label{tab:pgt-main}
\centering\scriptsize
\setlength{\tabcolsep}{5pt}
\renewcommand{\arraystretch}{1.2}

\begin{subtable}{\textwidth}
\centering
\caption{Joints: MPJPE / PA-MPJPE}
\resizebox{\linewidth}{!}{%
\begin{tabular}{l cc cc cc cc cc cc}
\toprule
& \multicolumn{2}{c}{H36M} & \multicolumn{2}{c}{MPI-3D} & \multicolumn{2}{c}{PoseTrack} & \multicolumn{2}{c}{CrowdPose} & \multicolumn{2}{c}{OCHuman} & \multicolumn{2}{c}{\textit{Average}} \\
\cmidrule(lr){2-3} \cmidrule(lr){4-5} \cmidrule(lr){6-7} \cmidrule(lr){8-9} \cmidrule(lr){10-11} \cmidrule(l){12-13}
Deterministic & 19.2 & 13.5 & 25.4 & 17.6 & 24.9 & 16.7 & 33.2 & 20.0 & 55.2 & 30.5 & 31.6 & 19.7 \\
Param-NLL & 22.6 & 13.2 & 27.9 & 17.0 & 25.4 & \textbf{15.5} & 31.3 & 17.8 & 53.1 & 28.8 & 32.0 & 18.5 \\
Matrix-Fisher & 26.2 & 15.0 & 33.7 & 19.9 & 32.5 & 18.8 & 35.3 & 19.8 & 51.6 & 28.9 & 35.9 & 20.5 \\
Normalizing Flow & 22.3 & 13.3 & 30.1 & 17.3 & 28.1 & 16.1 & 31.7 & \textbf{17.5} & 47.8 & 26.4 & 32.0 & 18.1 \\
\midrule
\rowcolor{gray!12}
\textbf{CLUE} & \textbf{18.2} & \textbf{12.7} & \textbf{24.8} & \textbf{16.2} & \textbf{24.6} & 15.7 & \textbf{29.1} & 17.7 & \textbf{40.5} & \textbf{26.1} & \textbf{27.5} & \textbf{17.7} \\
\bottomrule
\end{tabular}%
}
\end{subtable}

\vspace{6pt}
\begin{subtable}{\textwidth}
\centering
\caption{Surface: PVE / PA-PVE}
\resizebox{\linewidth}{!}{%
\begin{tabular}{l cc cc cc cc cc cc}
\toprule
& \multicolumn{2}{c}{H36M} & \multicolumn{2}{c}{MPI-3D} & \multicolumn{2}{c}{PoseTrack} & \multicolumn{2}{c}{CrowdPose} & \multicolumn{2}{c}{OCHuman} & \multicolumn{2}{c}{\textit{Average}} \\
\cmidrule(lr){2-3} \cmidrule(lr){4-5} \cmidrule(lr){6-7} \cmidrule(lr){8-9} \cmidrule(lr){10-11} \cmidrule(l){12-13}
Deterministic & 22.6 & 17.2 & 28.7 & 21.5 & 28.3 & 20.6 & 35.9 & 24.2 & 61.2 & 37.3 & 35.4 & 24.2 \\
Param-NLL & 25.5 & 16.3 & 30.2 & 20.1 & 28.1 & 18.9 & 33.5 & 21.5 & 57.6 & 35.0 & 35.0 & 22.3 \\
Matrix-Fisher & 28.4 & 18.3 & 35.9 & 23.0 & 34.6 & 21.6 & 37.8 & 23.8 & 57.7 & 35.5 & 38.9 & 24.4 \\
Normalizing Flow & 25.1 & 16.9 & 32.7 & 20.7 & 30.7 & 19.4 & 34.1 & \textbf{21.3} & 53.7 & \textbf{32.7} & 35.3 & 22.2 \\
\midrule
\rowcolor{gray!12}
\textbf{CLUE} & \textbf{20.6} & \textbf{15.6} & \textbf{27.0} & \textbf{19.4} & \textbf{26.8} & \textbf{18.8} & \textbf{31.1} & 21.5 & \textbf{46.4} & 33.2 & \textbf{30.4} & \textbf{21.7} \\
\bottomrule
\end{tabular}%
}
\end{subtable}

\vspace{6pt}
\begin{subtable}{\textwidth}
\centering
\caption{Uncertainty: vertex-NLL / $\sigma$--error correlation}
\resizebox{\linewidth}{!}{%
\begin{tabular}{l cc cc cc cc cc cc}
\toprule
& \multicolumn{2}{c}{H36M} & \multicolumn{2}{c}{MPI-3D} & \multicolumn{2}{c}{PoseTrack} & \multicolumn{2}{c}{CrowdPose} & \multicolumn{2}{c}{OCHuman} & \multicolumn{2}{c}{\textit{Average}} \\
\cmidrule(lr){2-3} \cmidrule(lr){4-5} \cmidrule(lr){6-7} \cmidrule(lr){8-9} \cmidrule(lr){10-11} \cmidrule(l){12-13}
Deterministic & -- & -- & -- & -- & -- & -- & -- & -- & -- & -- & -- & -- \\
Param-NLL & 2.10 & 0.68 & 2.30 & 0.71 & 2.22 & 0.69 & 2.32 & 0.69 & 2.83 & 0.66 & 2.35 & 0.68 \\
Matrix-Fisher & 2.58 & 0.36 & 2.71 & 0.51 & 2.70 & 0.61 & 2.73 & 0.62 & 2.96 & 0.66 & 2.74 & 0.56 \\
Normalizing Flow & 2.92 & 0.67 & 3.11 & \textbf{0.73} & 2.90 & 0.68 & 2.99 & 0.67 & 3.48 & 0.68 & 3.08 & 0.69 \\
\midrule
\rowcolor{gray!12}
\textbf{CLUE} & \textbf{1.82} & \textbf{0.71} & \textbf{2.00} & \textbf{0.72} & \textbf{2.07} & \textbf{0.74} & \textbf{2.15} & \textbf{0.77} & \textbf{2.61} & \textbf{0.79} & \textbf{2.13} & \textbf{0.75} \\
\bottomrule
\end{tabular}%
}
\end{subtable}

\vspace{6pt}
\begin{subtable}{\textwidth}
\centering
\caption{Uncertainty: AUSE}
\resizebox{0.8\linewidth}{!}{%
\begin{tabular}{l c c c c c c}
\toprule
& H36M & MPI-3D & PoseTrack & CrowdPose & OCHuman & \textit{Average} \\
\cmidrule(lr){2-2} \cmidrule(lr){3-3} \cmidrule(lr){4-4} \cmidrule(lr){5-5} \cmidrule(lr){6-6} \cmidrule(l){7-7}
Deterministic & -- & -- & -- & -- & -- & -- \\
Param-NLL & 0.081 & 0.072 & 0.087 & 0.072 & 0.080 & 0.078 \\
Matrix-Fisher & 0.159 & 0.142 & 0.117 & 0.117 & 0.111 & 0.129 \\
Normalizing Flow & 0.080 & 0.068 & 0.084 & 0.074 & 0.080 & 0.077 \\
\midrule
\rowcolor{gray!12}
\textbf{CLUE} & \textbf{0.065} & \textbf{0.067} & \textbf{0.079} & \textbf{0.066} & \textbf{0.059} & \textbf{0.067} \\
\bottomrule
\end{tabular}%
}
\end{subtable}
\end{table}